\documentclass{article}

\PassOptionsToPackage{numbers, compress}{natbib}
    \usepackage[final]{neurips_2021}

\usepackage[utf8]{inputenc} 
\usepackage[T1]{fontenc}    
\usepackage{hyperref}       
\usepackage{url}            
\usepackage{booktabs}       
\usepackage{amsfonts}       
\usepackage{nicefrac}       
\usepackage{microtype}      
\usepackage{xcolor}         
\usepackage{enumitem}
\usepackage{float}
\usepackage[ruled,vlined,noend]{algorithm2e} 
\usepackage{etoolbox}
\makeatletter
\patchcmd{\@algocf@start}
  {-1.5em}
  {0pt}
  {}{}
\makeatother
\usepackage{multicol}
\usepackage{svg}

\usepackage{hyperref}
\usepackage{url}
\usepackage{bm}
\usepackage{amsmath}
\usepackage{subcaption}
\usepackage{tikz}
\usepackage{arydshln}
\usepackage[symbol]{footmisc}
\usepackage{footnote}
\usepackage{multirow}
\usepackage[T1]{fontenc}    
\usepackage{graphicx}
\usepackage{wrapfig}
\usepackage{wrapfig,lipsum,booktabs}
\usepackage{amssymb}
\usepackage{array}
\usepackage{arydshln}
\usepackage{setspace}
\usepackage{comment}

\hypersetup{colorlinks,citecolor={teal}}

\usepackage{caption}
\usepackage{subcaption}

\usepackage{amsmath,amsfonts}
\usepackage{amsthm}
\usepackage{amssymb,amsopn}
\usepackage{bm} 

\usepackage{graphicx}  

\DeclareMathOperator*{\argmax}{arg\,max}

\newlength{\widebarargwidth}
\newlength{\widebarargheight}
\newlength{\widebarargdepth}

\newcommand{\eat}[1]{}

\usepackage{xcolor}
\definecolor{navyblue}{rgb}{0.0, 0.0, 0.5}
\definecolor{darkblue}{rgb}{0.0, 0.0, 0.55}
\hypersetup{
    colorlinks = true,
    citecolor=darkblue,
    linkcolor=red,
    urlcolor  = blue,
    anchorcolor = blue}

\usepackage{supertabular}
\usepackage{array,booktabs}
\usepackage{amsmath}
\usepackage{bm}
\usepackage{amsfonts}
\usepackage{multirow}

\usepackage{longtable}

\newcounter{daggerfootnote}
\newcommand*{\daggerfootnote}[1]{%
    \setcounter{daggerfootnote}{\value{footnote}}%
    \renewcommand*{\thefootnote}{\fnsymbol{footnote}}%
    \footnote[2]{#1}%
    \setcounter{footnote}{\value{daggerfootnote}}%
    \renewcommand*{\thefootnote}{\arabic{footnote}}%
    }

\title{Task-Adaptive Neural Network Search with Meta-Contrastive Learning}

\author{%
  Wonyong Jeong$^{1,2}$\thanks{Equal contribution.}~~\thanks{This work was done while the author was interning at AITRICS.} \hspace{0.05in} 
  Hayeon Lee$^{1,2}$\footnotemark[1]~~\footnotemark[2] \hspace{0.05in} 
  Geon Park$^{1,2}$\footnotemark[1]~~\footnotemark[2] \hspace{0.05in} \\ 
  \textbf{Eunyoung Hyung$^{1,2}$\footnotemark[2] \hspace{0.05in}
  Jinheon Baek$^{1}$ \hspace{0.05in} 
  Sung Ju Hwang$^{1, 2}$}\\
  
  KAIST$^{1}$, AITRICS$^{2}$, Seoul, South Korea \\
  \texttt{wyjeong@kaist.ac.kr, hayeon926@kaist.ac.kr, geon.park@kaist.ac.kr} \\ 
  \texttt{ey0301.hyung@samsung.com, jinheon.baek@kaist.ac.kr, sjhwang82@kaist.ac.kr} \\
}

\begin{document}

\maketitle

\begin{abstract}
    Most conventional Neural Architecture Search (NAS) approaches are limited in that they only generate architectures without searching for the optimal parameters. While some NAS methods handle this issue by utilizing a supernet trained on a large-scale dataset such as ImageNet, they may be suboptimal if the target tasks are highly dissimilar from the dataset the supernet is trained on. To address such limitations, we introduce a novel problem of \emph{Neural Network Search} (NNS), whose goal is to search for the optimal pretrained network for a novel dataset and constraints (e.g. number of parameters), from a model zoo. Then, we propose a novel framework to tackle the problem, namely \emph{Task-Adaptive Neural Network Search} (TANS). Given a model-zoo that consists of network pretrained on diverse datasets, we use a novel amortized meta-learning framework to learn a cross-modal latent space with contrastive loss, to maximize the similarity between a dataset and a high-performing network on it, and minimize the similarity between irrelevant dataset-network pairs. We validate the effectiveness and efficiency of our method on ten real-world datasets, against existing NAS/AutoML baselines. The results show that our method instantly retrieves networks that outperform models obtained with the baselines with significantly fewer training steps to reach the target performance, thus minimizing the total cost of obtaining a task-optimal network. Our code and the model-zoo are available at  \href{https://github.com/wyjeong/TANS}{https://github.com/wyjeong/TANS}.
    
\end{abstract}

\section{Introduction}
\label{introduction}

\emph{Neural Architecture Search} (NAS) aims to automate the design process of network architectures by searching for high-performing architectures with RL~\cite{zoph2017rl, zoph2018learning}, evolutionary algorithms~\cite{real2019regularized, chen2019evolve}, parameter sharing~\cite{brock2018smash, pham2918enas}, or surrogate schemes~\cite{luo2018neural}, to overcome the excessive cost of trial-and-error approaches with the manual design of neural architectures~\cite{simonyan2014vgg, he2016resnet, howard2017mobilenet}. Despite their success, existing NAS methods suffer from several limitations, which hinder their applicability to practical scenarios. First of all, the search for the optimal architectures usually requires a large amount of computation, which can take multiple GPU hours or even days to finish. This excessive computation cost makes it difficult to efficiently obtain an optimal architecture for a novel dataset. Secondly, most NAS approaches only search for optimal architectures, without the consideration of their parameter values. Thus, they require extra computations and time for training on the new task, in addition to the architecture search cost, which is already excessively high. 

\begin{figure*}[t]
    \small
    \centering
    \includegraphics[width=1.0\textwidth]{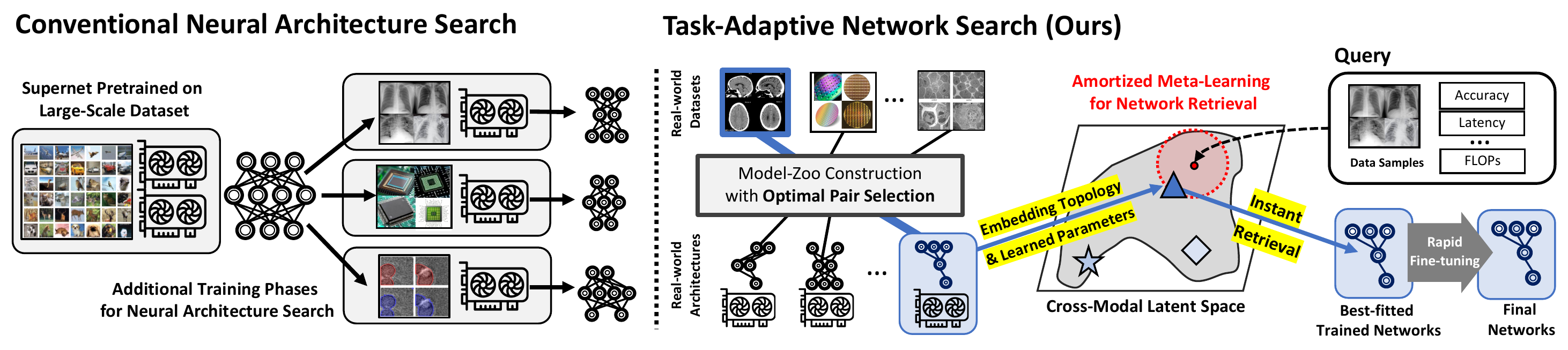} 
    \vspace{-0.25in}
    \caption{\small \textbf{Comparison between conventional NAS and our method}: Conventional supernet-based NAS approaches (Left) sample subnets from a fixed supernet trained on a single dataset. TANS (Right) can dynamically select the best-fitted neural networks that are trained on diverse datasets, adaptively for each query dataset.}
    \label{fig:overview}
    \vspace{-0.25in}
\end{figure*}

\vspace{-0.025in}
For this reason, supernet-based methods~\cite{cai2020once, lu2020nsganetv2} that search for a sub-network (subnet) from a network pretrained on large-scale data, are attracting more popularity as it eliminates the need for additional training. However, this approach may be suboptimal when we want to find the subnet for a dataset that is largely different from the source dataset the supernet is trained on (e.g. medical images or defect detection for semiconductors). This is a common limitation of existing NAS approaches, although the problem did not receive much attention due to the consideration of only a few datasets in the NAS benchmarks (See Figure~\ref{fig:overview}, left). However, in real-world scenarios, NAS approaches should search over diverse datasets with heterogeneous distributions, and thus it is important to \emph{task-adaptively} search for the architecture and parameter values for a given dataset. Recently, MetaD2A~\cite{lee2021rapid} has utilized meta-learning to learn common knowledge for NAS across tasks, to rapidly adapt to unseen tasks. However it does not consider parameters for the searched architecture, and thus still requires additional training on unseen datasets.

\vspace{-0.025in}
Given such limitations of NAS and meta-NAS methods, we introduce a novel problem of \emph{Neural Network Search} (NNS), whose goal is to search for the optimal pretrained networks for a given dataset and conditions (e.g. number of parameters). To tackle the problem, we propose a novel and extremely efficient task-adaptive neural network retrieval framework that searches for the optimal neural network with both the architecture and the parameters for a given task, based on cross-modal retrieval. In other words, instead of searching for an optimal architecture from scratch or taking a sub-network from a single super-net, we \emph{retrieve} the most optimal network for a given dataset in a task-adaptive manner (See Figure~\ref{fig:overview}, right), by searching through the model zoo that contains neural networks pretrained on diverse datasets. We first start with the construction of the model zoo, by pretraining state-of-the-art architectures on diverse real-world datasets. 

\vspace{-0.025in}
\begin{wrapfigure}[9]{r}{0.3\textwidth}
    \vspace{-0.35in}
    \small
    \centering
    \includegraphics[width=0.285\textwidth]{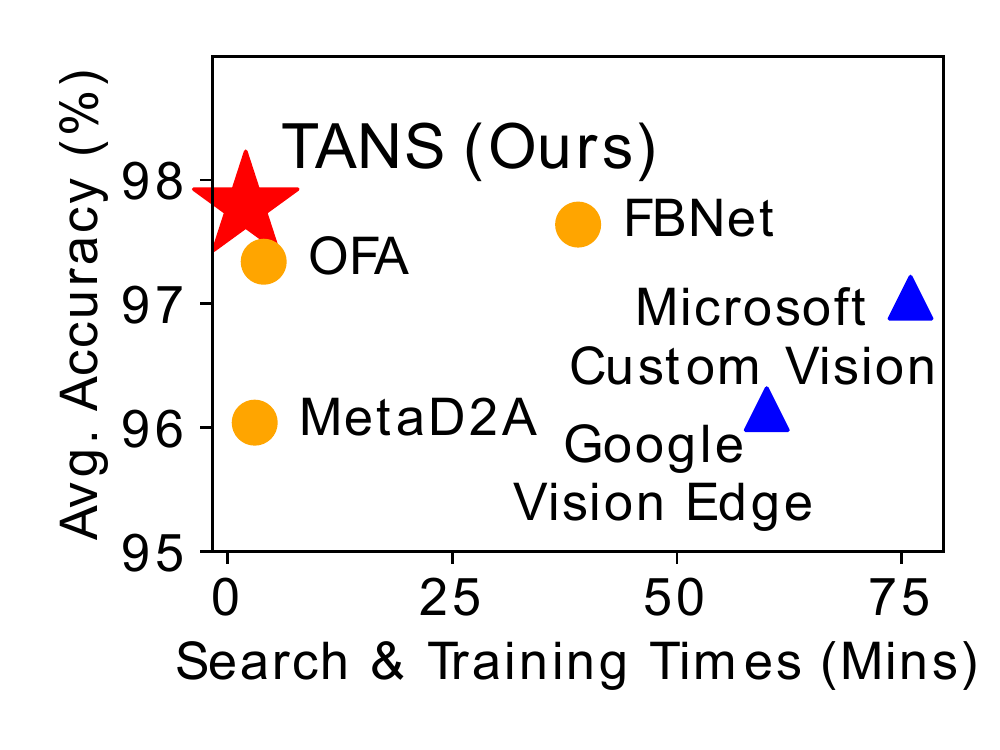}
    \vspace{-0.175in}
    \caption{\small Comparison with NAS (orange) \& AutoML (blue) baselines on 5 Real-world Datasets}
    \label{fig:automl}
\end{wrapfigure} 
Then, we train our retrieval model via amortized meta-learning of a cross-modal latent space with a contrastive learning objective. Specifically, we encode each dataset with a set encoder and obtain functional and topological embeddings of a network, such that a dataset is embedded closer to the network that performs well on it while minimizing the similarity between irrelevant dataset-network pairs. The learning process is further guided by a performance predictor, which predicts the model's performance on a given dataset. 

\vspace{-0.025in}
The proposed \emph{Task-Adaptive Network Search} (TANS) largely outperforms conventional NAS/AutoML methods (See Figure~\ref{fig:automl}), while significantly reducing the search time. This is because the retrieval of a trained network can be done instantly without any additional architecture search cost, and retrieving a task-relevant network will further reduce the fine-tuning cost. To evaluate the proposed TANS, we first demonstrate the sample-efficiency of our model zoo construction method, over construction with random sampling of dataset-network pairs. Then, we show that the TANS can adaptively retrieve the best-fitted models for an unseen dataset. Finally, we show that our method significantly outperforms baseline NAS/AutoML methods on \textbf{real-world datasets} (Figure~\ref{fig:automl}), with incomparably smaller computational cost to reach the target performance. In sum, our main contributions are as follows:

\vspace{-0.05in}
\begin{itemize}[itemsep=0.1mm] 
    \vspace{-0.05in}
    \item We consider a novel problem of \emph{Neural Network Search}, whose goal is to search for the optimal network for a given task, including both the architecture and the parameters. 
    \vspace{-0.05in}
	\item We propose a novel cross-modal retrieval framework to retrieve a pretrained network from the model zoo for a given task via amortized meta-learning with constrastive objective. 
	\vspace{-0.05in}
    \item We propose an efficient model-zoo construction method to construct an effective database of dataset-architecture pairs, considering both the model performance and task diversity. 
	\vspace{-0.05in}
	\item We train and validate TANS on a newly collected large-scale database, on which our method outperforms all NAS/AutoML baselines with almost no architecture search cost and significantly fewer fine-tuning steps. 
\end{itemize}


	
\section{Related Work}
\label{related_work}
\paragraph{Neural Architecture Search} \emph{Neural Architecture Search} (NAS), which aims to automate the design of neural architectures, is an active topic of research nowadays. Earlier NAS methods use non-differentiable search techniques based on RL~\citep{zoph2017rl, zoph2018learning} or evolutionary algorithms~\cite{real2019regularized, chen2019evolve}. However, their excessive computational requirements~\citep{ren2020survey} in the search process limits their practical applicability to resource-limited settings. To tackle this challenge, one-shot methods share the parameters~\cite{pham2918enas, brock2018smash, liu2018darts, xu2020pc} among architectures, which reduces the search cost by orders of magnitude. The surrogate scheme predicts the performance of architectures without directly training them~\citep{luo2018neural, zhou2020econas, tang2020semi}, which also cuts down the search cost. Latent space-based NAS methods~\citep{luo2018neural, tang2020semi, yan2020does} learn latent embeddings of the architectures to reconstruct them for a specific task. Recently, supernet-based approaches, such as OFA~\cite{cai2020once}, receive the most attention due to their high-performances. OFA generates the subnet with its parameters by splitting the trained supernet. While this eliminates the need for costly re-training of each searched architecture from scratch, but, it only trains a fixed-supernet on a single dataset (ImageNet-1K), which limits their effectiveness on diverse tasks that are largely different from the training set. Whereas our TANS task-adaptively retrieves a trained neural network from a database of networks with varying architectures trained on diverse datasets.

\vspace{-0.1in}
\paragraph{Meta-Learning} The goal of meta-learning~\cite{thrun98} is to learn a model to generalize over the distribution of tasks, instead of instances from a single task, such that a meta-learner trained across multiple tasks can rapidly adapt to a novel task. While most meta-learning methods consider few-shot classification with a fixed architecture~\citep{vinyals2016matching, finn2017model, snell2017prototypical, nichol2018first, lee2019meta, lee2020l2b}, there are a few recent studies that couple NAS with meta-learning~\citep{shaw2019metanas1, lian2020metanas2, elsken2020metanas3} to search for the well-fitted architecture for the given task. However, these NAS approaches are limited to small-scale tasks due to the cost of roll-out gradient steps. To tackle this issue, MetaD2A~\cite{lee2021rapid} proposes to generate task-dependent architectures with amortized meta-learning, but does not consider parameters for the searched architecture, and thus requires additional cost of training it on unseen datasets. To overcome these limitations, our method retrieves the best-fitted architecture with its parameters for the target task, by learning a cross-modal latent space for dataset-network pairs with amortized meta-learning.

\vspace{-0.1in}
\paragraph{Neural Retrieval} Neural retrieval aims to search for and return the best-fitted item for the given query, by learning to embed items in a latent space with a neural network. Such approaches can be broadly classified into models for image retrieval~\citep{gordo2017end, devaraj2020efficient, yan2020deep} and text retrieval~\citep{zhang2020deep, chang2020ir1, xiong2021ir2}. Cross-modal retrieval approaches~\citep{lee2018crossmodal2, zhen2019crossmodal1, wang2020crossmodal3} handle retrieval across different modalities of data (e.g. image and text), by learning a common representation space to measure the similarity across the instances from different modalities. To our knowledge, none of the existing works is directly related to our approach that performs cross-modal retrieval of neural networks given datasets.

\section{Methodology}
\vspace{-0.05in}
We first define the Neural Network Search (NNS) problem and propose a meta-contrastive learning framework to learn a cross-modal retrieval space. We then describe the structural details of the query and model encoders, and an efficient model-zoo construction strategy. 

\vspace{-0.1in}
\subsection{Problem Definition}
\vspace{-0.05in}
Our goal in NNS is to search for an optimal network (with both architectures and parameters) for a dataset and constraints, by learning a cross-modal latent space over the dataset-network pairs. We first describe the task-adaptive network retrieval with amortized meta-contrastive learning.

\subsubsection{Meta-Training}
\vspace{-0.05in}
We assume that we have a database of neural networks (model zoo) pretrained over a distribution of tasks $p(\tau)$ with each task $\tau= \{ D^\tau, M^\tau, s^\tau \}$, where $D^\tau$ denotes a dataset, $M^\tau$ denotes a neural network (or model) trained on the dataset, and $s^\tau$ denotes a set of auxiliary constraints for the network to be found (e.g. number of parameters or the inference latency). Then, our goal is to learn a cross-modal latent space for the dataset-network pairs $(D^\tau, M^\tau)$ while considering the constraints $s^\tau$ over the task distribution $p({\tau})$, as illustrated in Figure~\ref{fig:overview}. In this way, a meta-trained model from diverse $(D^\tau, M^\tau)$ pairs, will rapidly generalize on an unseen dataset $\tilde{D}$; $D^\tau \cap \tilde{D} = \emptyset$ by retrieving a well-fitted neural network on the dataset that satisfies the conditions $s^\tau$.

\begin{figure*}[t]
\small
    \centering
    \includegraphics[width=0.95\textwidth]{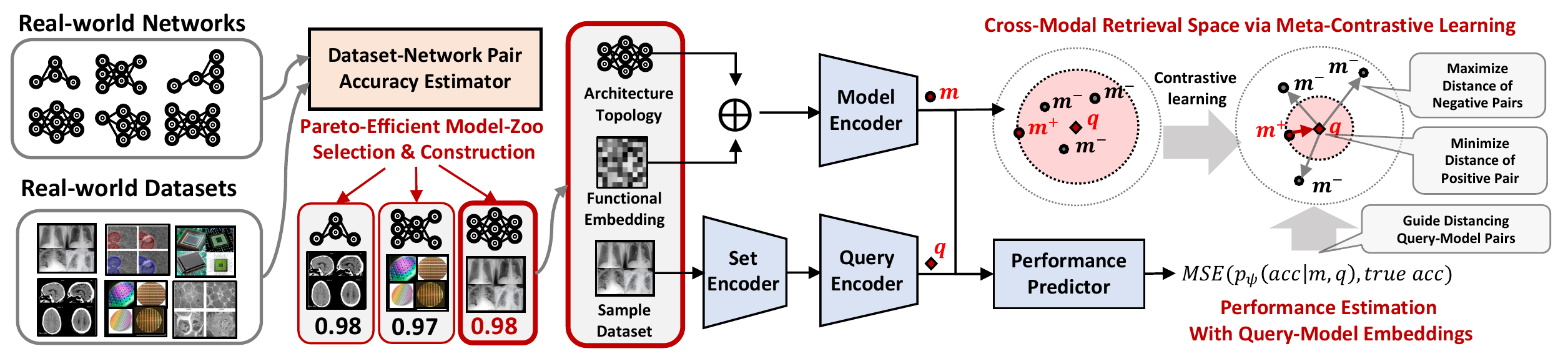}
    \vspace{-0.05in}
    \caption{\small \textbf{Illustration for overall framework of our proposed method (TANS)}: We first construct our model-zoo with pareto-optimal pairs of dataset and network, rather than exhaustively train all possible pairs. We then embed a model and a dataset with a graph-functional model and a set encoder. After that, we meta-learn the cross-modal retrieval network over multiple model-query pairs, guided by our performance predictor.}
    \label{fig:hn}
    \vspace{-0.1in}
\end{figure*}

\vspace{-0.1in}
\paragraph{Task-Adaptive Neural Network Retrieval}
To learn a cross-modal latent space for dataset-network pairs over a task distribution, we first introduce a novel task-adaptive neural network retrieval problem. The goal of task-adaptive retrieval is to find an appropriate network $M^\tau$ given the query dataset $D^\tau$ for task $\tau$. To this end, we need to calculate the similarity between the dataset-network pair $(D^\tau, M^\tau) \in \mathcal{Q} \times \mathcal{M}$, with a scoring function $f$ that outputs the similarity between them as follows:
\begin{equation}
    \begin{aligned}
    &\max_{\bm{\theta}, \bm{\phi}} \sum_{\tau \in p(\tau)} f(\bm{q}, \bm{m}), 
    &\bm{q} =  E_Q(D^\tau; \bm{\theta}) \quad \text{and} \quad \bm{m} = E_M(M^\tau; \bm{\phi}),
    \label{eq:1}
    \end{aligned}
\end{equation}
where $E_Q: \mathcal{Q} \rightarrow \mathbb{R}^d$ is a query (dataset) encoder, $E_M: \mathcal{M} \rightarrow \mathbb{R}^d$ is a model encoder, which are parameterized with the parameter $\bm{\theta}$ and $\bm{\phi}$ respectively, and $f_{sim}: \mathbb{R}^d \times \mathbb{R}^d \rightarrow \mathbb{R}$ is a scoring function for the query-model pair. In this way, we can construct the cross-modal latent space for dataset-network pairs over the task distribution with equation \ref{eq:1}, and use this space to rapidly retrieve the well-fitted neural network in response to the unseen query dataset.

We can learn such a cross-modal latent space of dataset-network pairs for rapid retrieval by directly solving for the above objective, with the assumption that we have the query and the model encoder: $Q$ and $M$. However, we further propose a contrastive loss to maximize the similarity between a dataset and a network that obtains high performance on it in the learned latent space, and minimize the similarity between irrelevant dataset-network pairs, inspired by~\citet{Faghri18, Martin18}. While existing works such as~\citet{Faghri18, Martin18} target image-to-text retrieval, we tackle the problem of cross-modal retrieval across datasets and networks, which is a nontrivial problem as it requires task-level meta-learning.

\vspace{-0.05in}
\paragraph{Retrieval with Meta-Contrastive Learning}
Our \emph{meta-contrastive learning} objective for each task $\tau \in p(\tau)$ consisting of a dataset-model pair $(D^\tau, M^\tau) \in \mathcal{Q} \times \mathcal{M}$, aims to maximize the similarity between positive pairs: $f_{sim}(\bm{q}, \bm{m}^+)$, while minimizing the similarity between negative pairs: $f_{sim}(\bm{q}, \bm{m}^-)$, where $\bm{m}^+$ is obtained from the sampled target task $\tau \in p(\tau)$ and $\bm{m}^-$ is obtained from other tasks $\gamma \in p(\tau)$; $\gamma \neq \tau$, which is illustrated in Figure~\ref{fig:hn}. This meta-contrastive learning loss can be formally defined as follows:
\begin{align}
    &\mathcal{L}_{m}(\tau; \bm{\theta}, \bm{\phi})=\mathcal{L}(f_{sim}(\bm{q}, \bm{m}^+), f_{sim}(\bm{q}, \bm{m}^-); \bm{\theta}, \bm{\phi}) \\
    &\bm{q} =  E_Q(D^\tau; \bm{\theta}), \; \bm{m^+} = E_M(M^\tau; \bm{\phi}), \; \bm{m^-} = E_M(M^\gamma; \bm{\phi}) \nonumber.
\end{align}

We then introduce $\mathcal{L}$ for the meta-contrastive learning:
\begin{equation}
    \max \left( 0, \alpha -\log \frac{\exp(f_{sim}(\bm{q}, \bm{m}^+))}{\exp \left( \sum_{\gamma \in p(\tau), \gamma \neq \tau} f_{sim}(\bm{q}, \bm{m}^-) \right) } \right),
\end{equation}
where $\alpha \in \mathbb{R}$ is a margin hyper-parameter and the score function $f_{sim}$ is the cosine similarity. The contrastive loss promotes the positive $(\bm{q}, \bm{m}^+)$ embedding pair to be close together, with at most margin $\alpha$ closer than the negative $(\bm{q}, \bm{m}^-)$ embedding pairs in the learned cross-modal metric space. Note that, similar to this, we also contrast the query with its corresponding model: $\mathcal{L}_{q}(f_{sim}(\bm{q}^+, \bm{m}), f_{sim}(\bm{q}^-, \bm{m}))$, which we describe in the supplementary material in detail.



With the above ingredients, we minimize the meta-contrastive learning loss over a task distribution $p(\tau)$, defined with the model $\mathcal{L}_{m}$ and query $\mathcal{L}_{q}$ contrastive losses, as follows:
\begin{equation}
    \min_{\bm{\phi}, \bm{\theta}} \sum_{\tau \in p (\tau)} \mathcal{L}_{m}(\tau; \bm{\theta}, \bm{\phi}) + \mathcal{L}_{q}(\tau; \bm{\theta}, \bm{\phi}).
\label{eq:objective}
\end{equation}

\vspace{-0.1in}
\paragraph{Meta-Performance Surrogate Model} 
We propose the meta-performance surrogate model to predict the performance on an unseen dataset without directly training on it, which is highly practical in real-world scenarios since it is expensive to iteratively train models for every dataset to measure the performance. Thus, we meta-train a performance surrogate model $a=S(\tau; {\bm{\psi}})$ over a distribution of tasks $p({\tau})$ on the model-zoo database. This model not only accurately predicts the performance $a$ of a network $M^\tau$ on an unseen dataset $D^\tau$, but also guides the learning of the cross-modal retrieval space, thus embedding a neural network closer to datasets that it performs well on. 

Specifically, the proposed surrogate model $S$ takes a query embedding $\bm{q}^\tau$ and a model embedding $\bm{m}^\tau$ as an input for the given task $\tau$, and then forwards them to predict the accuracy of the model for the query. We train this performance predictor $S(\tau;\bm{\psi})$ to minimize the mean-square error loss $\mathcal{L}_{s}(\tau;\bm{\psi}) = (s^\tau_{\text{acc}} - S(\tau; {\bm{\psi}}))^2$ between the predicted accuracy $S(\tau; {\bm{\psi}})$ and the true accuracy $s^\tau_{\text{acc}}$ for the model on each task $\tau$, which is sampled from the task distribution $p({\tau})$. Then, we combine this objective with retrieval objective in equation~\ref{eq:objective} to train the entire framework as follows:
\begin{equation}
    \min_{\bm{\phi}, \bm{\theta}, \bm{\psi}} \sum_{\tau \in p (\tau)} \mathcal{L}_{m}(\tau; \bm{\theta}, \bm{\phi}) + \mathcal{L}_{q}(\tau; \bm{\theta}, \bm{\phi}) + \lambda \cdot \mathcal{L}_{s}(\tau;\bm{\psi}),
\label{eq:final_objective}
\end{equation}
where $\lambda$ is a hyper-parameter for weighting losses.

\subsubsection{Meta-Test}
By leveraging the meta-learned cross-modal retrieval space, we can instantly retrieve the best-fitted pretrained network $M \in \mathcal{M}$, given an unseen query dataset $\tilde{D} \in \tilde{\mathcal{Q}}$, which is disjoint from the meta-training dataset $D \in \mathcal{Q}$. Equipped with meta-training components described in the previous subsection, we now describe the details of our model at inference time, which includes the following: amortized inference, performance prediction, and task- and constraints-adaptive initialization.


\vspace{-0.05in}
\paragraph{Amortized Inference} Most existing NAS methods are slow as they require several GPU hours for training, to find the optimal architecture for a dataset $\tilde{D}$. Contrarily, the proposed \emph{Task-Adaptive Network Search} (TANS) only requires a single forward pass per dataset, to obtain a query embedding $\tilde{\bm{q}}$ for the unseen dataset using the query encoder $Q(\tilde{D};\bm{\theta}^*)$ with the meta-trained parameters $\bm{\theta}^*$, since we train our model with amortized meta-learning over a distribution of tasks $p(\tau)$. After obtaining the query embedding, we retrieve the best-fitted network $M^*$ for the query based on the similarity:
\begin{equation}
   M^* = \max_{M^\tau} \{f_{sim}(\tilde{\bm{q}}, \bm{m}^{\tau}) \mid \tau \in p(\tau)\},
\end{equation}
where a set of model embeddings $\{{\bm{m}^{\tau}} \mid \tau \in p(\tau)\}$ is pre-computed by the meta-trained model encoder $E_M(M^{\tau};\bm{\phi}^*)$.


 

\vspace{-0.05in}
\paragraph{Performance Prediction} While we achieve the competitive performance on unseen dataset only with the previously defined model, we also use the meta-learned performance predictor $S$ to select the best performing one among top $K$ candidate networks $\{\tilde{M}_i\}_{i=1}^{K}$ based on their predicted performances. Since this surrogate model with module to consider datasets is meta-learned over the distribution of tasks $p(\tau)$, we predict the performance on an unseen dataset $\tilde{D}$ without training on it. This is different from the conventional surrogate models~\cite{luo2018neural, zhou2020econas, tang2020semi} that additionally need to train on an unseen dataset from scratch, to predict the performance on it. 


\vspace{-0.05in}
\paragraph{Task-adaptive Initialization} Given an unseen dataset, the proposed TANS can retrieve the network trained on a training dataset that is highly similar to the unseen query dataset from the model zoo (See Figure~\ref{fig:unseen_retrieval}). Therefore, fine-tuning time of the trained network for the unseen target dataset $\tilde{D}$ is effectively reduced since the retrieved network $M$ has task-relevant initial parameters that are already trained on a similar dataset. If we need to further consider constraints $s$, such as parameters and FLOPs, then we can easily check if the retrieved models meet the specific constraints by sorting them in the descending order of their scores, and then selecting the constraint-fitted best accuracy model.


\subsection{Encoding Datasets and Networks}
\vspace{-0.05in}
\label{sec:structure_details}


\paragraph{Query Encoder}
The goal of the proposed query encoder $E_Q(D;{\bm{\theta}}): \mathcal{Q} \rightarrow \mathbb{R}^d$ is to embed a dataset $D$ as a single query vector $\bm{q}$ onto the cross-modal latent space. Since each dataset $D$ consists of $n$ data instances $D = \left\{ X_i \right\}_{i=1}^n \in \mathcal{Q}$, we need to fulfill the permutation-invariance condition over the data instances $X_i$, to output a consistent representation regardless of the order of its instances. To satisfy this condition, we first individually transform $n$ randomly sampled instances for the dataset $D$ with a continuous learnable function $\rho$, and then apply a pooling operation to obtain the query vector $\bm{q} = \sum_{X_i \in D} \rho(X_i)$, adopting~\citet{deepset}.

\vspace{-0.05in}
\paragraph{Model Encoder}
To encode a neural network $M^\tau$, we consider both its architecture and the model parameters trained on the dataset $D^\tau$ for each task $\tau$. Thus, we propose to generate a model embedding with two encoding functions: 1) topological encoding and 2) functional encoding. 

Following~\citet{cai2020once}, we first obtain a topological embedding $\bm{v}^\tau_t$ with auxiliary information about the architecture topology, such as the numbers of layers, channel expansion ratios, and kernel sizes. Then, our next goal is to encode the trained model parameters for the given task, to further consider parameters on the neural architecture. However, a major problem here is that directly encoding millions of parameters into a vector is highly challenging and inefficient. To this end, we use functional embedding, which embeds a network solely based on its input-output pairs. This operation generates the embedding of trained networks, by feeding a fixed Gaussian random noise into each network $M^\tau$, and then obtaining an output $\bm{v}^\tau_f$ of it. The intuition behind the functional embedding is straightforward: since networks with different architectures and parameters comprise different functions, we can produce different outputs for the same input.

With the two encoding functions, the proposed model encoder generates the model representation by concatenating the topology and function embeddings $[\bm{v}^\tau_t, \bm{v}^\tau_f]$, and then transforming the concatenated vector with a non-linear function $\sigma$ as follows: $\bm{m}^\tau = \sigma([\bm{v}^\tau_t, \bm{v}^\tau_f]) = E_M(M^\tau;\bm{\phi})$. 
Note that, the two encoding functions satisfy the injectiveness property under certain conditions, which helps with the accurate retrieval of the embedded elements in a condensed continuous latent space. We provide the proof of the injectiveness of the two encoding functions in Section C of the supplementary file. 

\vspace{-0.05in}
\subsection{Model-Zoo Construction}
\label{sec:modelzoocons}
\vspace{-0.05in}
Given a set of datasets $\mathcal{D}=\{D_1, \dots, D_K\}$ and a set of architectures $\mathcal{M}=\{M_1, \dots, M_N\}$, the most straightforward way to construct a model zoo $\mathcal{Z}$, is by training all architectures on all datasets, which will yield a model zoo $\mathcal{Z}$ that contains $N\times K$ pretrained networks. However, we may further reduce the construction cost by collecting $P$ dataset-model pairs, $\{D,M\}_{i=1}^P$, where $P \ll N\times{K}$, by randomly sampling an architecture $M\in\mathcal{M}$ and then training it on $D\in\mathcal{D}$.
Although this works well in practice (see Figure~\ref{fig:model_zoo} (Bottom)), we further propose an efficient algorithm to construct it in a more sample-efficient manner, by skipping evaluation of dataset-model pairs that are certainly worse than others in all aspects (memory consumption, latency, and test accuracy).
We start with an initial model zoo $\mathcal{Z}_{(0)}$ that contains a small amount of randomly selected pairs and its test accuracies. Then, at each iteration $t$, among the set of candidates $C_{(t)}$, we find a pair $\{D, M\}$ that can expand the currently known set of all Pareto-optimal pairs w.r.t. all conditions (memory, latency, and test accuracy on the dataset $D$), based on the amount of the Pareto front expansion estimated by $f_{zoo}(\cdot; \mathcal{Z}_{(t)})$:
\begin{align}
   \{D, M\}_{(t+1)} &= \argmax_{(D, M) \in C_{(t)}} f_{zoo}(\{D,M\}; \mathcal{Z}_{(t)})
\end{align}
where $f_{zoo}(\{D,M\}; \mathcal{Z}) := \underset{\hat{s}_{acc}}{\mathbb{E}}[g_D(\mathcal{Z} \cup \lbrace D, M, \hat{s}_{acc} \rbrace ) - g_D(\mathcal{Z})]$, $\hat{s}_{acc} \sim S(\{D,M\}; {\bm{\psi}_{zoo}})$ with parameter $\bm{\psi}_{zoo}$ trained on $\mathcal{Z}$, and the function $g_D$ measures the volume under the Pareto curve, also known as the Hypervolume Indicator~\cite{nowak_empirical_2014}, for the dataset $D$. We then train $M$ on $D$, and add it to the current model zoo $\mathcal{Z}_{(t)}$. 
For the full algorithm, please refer to Appendix~\ref{suppl:implementation}.

\vspace{-0.05in}
\section{Experiments}
\label{experiments}
\vspace{-0.1in}
In this section, we conduct extensive experimental validation against conventional NAS methods and commercially available AutoML platforms, to demonstrate the effectiveness of our proposed method. 

\subsection{Experimental Setup}
\vspace{-0.05in}



\paragraph{Datasets} We collect \textbf{96} real-world image classification datasets from Kaggle\footnote[1]{\url{https://www.kaggle.com/}}. Then we divide the datasets into two non-overlapping sets for \textbf{86} meta-training and \textbf{10} meta-test datasets. As some datasets contain relatively large number of classes than the other datasets, we adjust each dataset to have up to 20 classes, yielding 140 and 10 datasets for meta-training and meta-test datasets, respectively (Please see Table~\ref{tbl:dataset} for detailed dataset configuration). For each dataset, we use randomly sampled 80/20\% instances as a training and test set. To be specific, our 10 meta-test datasets include Colorectal Histology, Drawing, Dessert, Chinese Characters, Speed Limit Signs, Alien vs Predator, COVID-19, Gemstones, and Dog Breeds. We strictly verified that there is \textbf{no dataset-, class-, and instance-level overlap} between the meta-training and the meta-test datasets, while some datasets may contain semantically similar classes.

\vspace{-0.05in}
\paragraph{Baseline Models} We consider MobileNetV3~\citep{howard2019searching} pretrained on ImageNet as our baseline neural network. We compare our method with conventional NAS methods, such as PC-DARTS~\citep{xu2020pc} and DrNAS~\citep{chen2021drnas}, weight-sharing approaches, such as FBNet~\citep{wu2019fbnet} and Once-For-All~\citep{cai2020once}, and data-driven meta-NAS approach, e.g. MetaD2A~\citep{lee2021rapid}. All these NAS baseline approaches are based on MobileNetV3 pretrained on ImageNet, except for the conventional NAS methods that are only able to generate architectures. As such conventional NAS methods start from the scratch, we train them for sufficient amount of training epochs (10 times more training steps) for fair comparison. Please see Appendix~\ref{suppl:train-details} for further detailed descriptions of the experimental setup.


\vspace{-0.05in}
\paragraph{Model-zoo Construction} We follow the OFA search space~\cite{cai2020once}, which allows us to design resource-efficient architectures, and thus we obtain network candidates from the OFA space. We sample 100 networks condidates per meta-training dataset, and then train the network-dataset pairs, yielding 14,000 dataset-network pairs in our full model-zoo. We also construct smaller-sized efficient model-zoos from the full model-zoo (14,000) with our efficient algorithm described in Section \ref{sec:modelzoocons}. We use the full-sized model-zoo as our base model-zoo, unless otherwise we clearly mention the size of the used model-zoo. Detailed descriptions, e.g. training details and costs, are described in Appendix~\ref{suppl:detail_model_zoo}.
 
\begin{table*}[t!]
	\small
	\vspace{-0.1in}
	\caption{\small \textbf{Performance of the searched networks on 10 unseen real-world datasets.} We report the averaged accuracy on ten unseen meta-test datasets over 3 different runs with $95\%$ confidence intervals.
	} 
	\vspace{-0.1in}
	\begin{center}
		\resizebox{\textwidth}{!}{
			\renewcommand{\arraystretch}{0.7}
			\begin{tabular}{clrccrrcc}
				\toprule
				\midrule
				\multirow{2}{*}{\textbf{Target Dataset}} &
				\multirow{2}{*}{\textbf{Method}} & 
				\multirow{2}{*}{\textbf{\# Epochs}} & 
				\textbf{FLOPs}&
				\textbf{Params} &
				\textbf{Search Time} &
				\textbf{Training Time} &
				\multirow{2}{*}{\textbf{Speed Up}} &
				\textbf{Accuracy} \\
				& & & (M) & (M) & (GPU sec) & (GPU sec) & & (\%) \\
				\midrule
				\midrule
				
				\multirow{8}{*}{\shortstack{ \\ \textbf{Averaged} \\ \textbf{Performance}}} &
				MobileNetV3~\citep{howard2019searching} &
				50 &
				132.94 &
				4.00 &
				- &
				257.78\tiny$\pm$09.77 &
				1.00$\times$ &
				94.20\tiny$\pm$0.70                   
				\\
				
				\cdashline{2-9}\noalign{\vskip 0.75ex}
				
				&
				PC-DARTS~\citep{xu2020pc} &
				500 &
				566.55 &
				\textbf{3.54} &
				1100.37\tiny$\pm$22.20 &
				5721.13\tiny$\pm$793.71 &
				0.04$\times$&
				79.22\tiny$\pm$1.69
				\\
				
				&
				DrNAS~\citep{chen2021drnas} &
				500 &
				623.43 &
				4.12 &
				1501.75\tiny$\pm$43.92&
				5659.77\tiny$\pm$403.62&
				0.04$\times$&
				84.06\tiny$\pm$0.97
				\\
				
				\cdashline{2-9}\noalign{\vskip 0.75ex}
				
				&
				FBNet-A~\citep{wu2019fbnet} &
				50 &
				246.69 &
				4.3 &
				- &
				293.42\tiny$\pm$57.45&
				0.88$\times$&
				93.00\tiny$\pm$1.95
				\\
				
				&
				OFA~\citep{cai2020once} &
				50 &
				148.76 &
				6.74 &
				121.90\tiny$\pm$0.00 &
				226.58\tiny$\pm$03.13 &
				0.74$\times$ &
				93.89\tiny$\pm$0.84               
				\\
				
				&
				MetaD2A~\citep{lee2021rapid} &
				50 &
				512.67 & 
				6.56 &
				2.59\tiny$\pm$0.13 &
				345.39\tiny$\pm$28.36 &
				0.74$\times$ &
				95.24\tiny$\pm$1.14                
				\\
				
				\cdashline{2-9}\noalign{\vskip 0.75ex}
				
				&
				\textbf{TANS (Ours) } &
				10 &
				181.74 &
				5.51 &
				0.002\tiny$\pm$0.00 &
				40.19\tiny$\pm$03.06 &
				- &
				95.17\tiny$\pm$2.20 
				\\
				
				&
				\textbf{TANS (Ours) } &
				50 &
				181.74 &
				5.51 &
				\textbf{0.002\tiny$\pm$0.00} &
				\textbf{200.93\tiny$\pm$11.01} &
				\textbf{1.28$\times$} &
				\textbf{96.28\tiny$\pm$0.30}
				\\
				
				\midrule
				
				\multirow{8}{*}{\shortstack{\\ \\ \\ Colorectal \\Histology \\ Dataset \\ (Easy)}} &
				MobileNetV3~\citep{howard2019searching} &
				50 &
				132.94 &
				\textbf{4.00} &
				- &
				577.18\tiny$\pm$04.15 &
				1.00$\times$ &
				96.23\tiny$\pm$0.07          
				\\
				
				\cdashline{2-9}\noalign{\vskip 0.75ex}
				
				&
				PC-DARTS~\citep{xu2020pc} &
				500 & 
				534.64 &
				4.02&
				2062.42\tiny$\pm$49.14&
				12124.18\tiny$\pm$1051.16&
				0.04$\times$&
				96.17\tiny$\pm$0.68
				\\
				
				&
				DrNAS~\citep{chen2021drnas} &
				500 &
				614.23 &
				4.12&
				4183.20\tiny$\pm$188.60&
				11355.18\tiny$\pm$1352.62&
				0.04$\times$&
				97.51\tiny$\pm$0.13
				\\
				
				\cdashline{2-9}\noalign{\vskip 0.75ex}
				&
				FBNet-A~\citep{wu2019fbnet} &
				50 &
				215.45 &
				4.3&
				-&
				696.00\tiny$\pm$295.19&
				0.83$\times$&
				95.43\tiny$\pm$0.57
				\\
				
				&
				OFA~\citep{cai2020once} &
				50 &
				134.85 &
				6.74 &
				121.90\tiny$\pm$0.00 &
				537.61\tiny$\pm$03.52 &
				0.88$\times$ &
				96.40\tiny$\pm$0.52                     
				\\
				
				&
				MetaD2A~\citep{lee2021rapid} &
				50 &
				506.88 &
				5.93 &
				2.58\tiny$\pm$0.12 &
				784.45\tiny$\pm$79.32 &
				0.73$\times$ &
				96.57\tiny$\pm$0.56 \\
				
				\cdashline{2-9}\noalign{\vskip 0.75ex}
				
				&
				\textbf{TANS (Ours)} &
				10 &
				171.74 &
				4.95 &
				0.001\tiny$\pm$0.00 &
				98.56\tiny$\pm$04.24 &
				- & 
				96.87\tiny$\pm$0.21                             
				\\
				
				&
				\textbf{TANS (Ours)} & 
				50 &
				171.74 &
				4.95 &
				\textbf{0.001\tiny$\pm$0.00} &
				\textbf{492.81\tiny$\pm$21.19} &
				\textbf{1.17$\times$} &
				\textbf{97.67\tiny$\pm$0.05}
				\\
				
				\midrule
				
				
				\multirow{8}{*}{\shortstack{\\ \\ \\ Food \\Classification\\ Dataset \\ (Hard)}} &
				MobileNetV3~\citep{howard2019searching} &
				50 &
				132.94 &
				4.00 &
				- &
				235.57\tiny$\pm$07.57 &
				1.00$\times$ &
				87.52\tiny$\pm$0.78  
				\\
				
				\cdashline{2-9}\noalign{\vskip 0.75ex}
				
				&
				PC-DARTS~\citep{xu2020pc} &
				500 &
				567.85 &
				\textbf{3.62}&
				1018.49\tiny$\pm$6.31&
				6323.40\tiny$\pm$938.83&
				0.03$\times$&
				55.42\tiny$\pm$2.46
				\\
				
				&
				DrNAS~\citep{chen2021drnas} &
				500 &
				632.67 &
				4.12&
				1276.38\tiny$\pm$0.00&
				5079.89\tiny$\pm$161.05&
				0.04$\times$&
				61.45\tiny$\pm$0.68
				\\
				
				\cdashline{2-9}\noalign{\vskip 0.75ex}
				
				&
				FBNet-A~\citep{wu2019fbnet} &
				50 &
				251.29 &
				4.3&
				-&
				251.24\tiny$\pm$3.31&
				0.94$\times$&
				84.33\tiny$\pm$1.41
				\\
				
				&
				OFA~\citep{cai2020once} &
				50 &
				152.34 &
				6.74 &
				121.90\tiny$\pm$0.00 &
				\textbf{190.86\tiny$\pm$03.48} &
				0.75$\times$ &
				87.43\tiny$\pm$0.59             
				\\
				
				&
				MetaD2A~\citep{lee2021rapid} &
				50 &
				521.11 &
				8.23 &
				2.60\tiny$\pm$0.23 &
				324.62\tiny$\pm$34.97 &
				0.72$\times$ &
				89.72\tiny$\pm$1.53             
				\\
				
				\cdashline{2-9}\noalign{\vskip 0.75ex}
				
				&
				\textbf{TANS (Ours) } & 
				10 &
				179.83 &
				5.07 &
				0.002\tiny$\pm$0.00 &
				40.59\tiny$\pm$04.84 &
				- & 
				93.11\tiny$\pm$0.24   
				\\
				
				&
				\textbf{TANS (Ours)} & 
				50 &
				179.83 &
				5.07 &
				\textbf{0.002\tiny$\pm$0.00} &
				202.93\tiny$\pm$24.21 &
				\textbf{1.16}$\times$ &
				\textbf{93.71\tiny$\pm$0.24}
				\\
				
				\midrule
				\bottomrule
			\end{tabular}
		}
	\end{center}
	\vspace{-0.1in}
	\label{tbl:unseen_task}
	\small
	\vspace{-0.1in}
\end{table*}

\begin{figure*}[t!]
\centering
\small{
\hspace{-0.05in}
Query \hspace{0.3in} Retrieval 
\hspace{0.75in} 
Query \hspace{0.3in} Retrieval
\hspace{0.75in} 
Query \hspace{0.3in} Retrieval
}

\hspace{-0.11in}
\subfloat{
\includegraphics[width=.15\columnwidth]{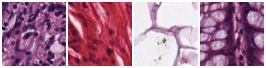}
\includegraphics[width=.15\columnwidth]{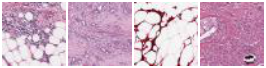}
}
\hspace{0.135in}
\subfloat{
\includegraphics[width=.15\columnwidth]{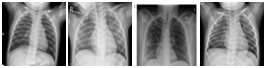}
\includegraphics[width=.15\columnwidth]{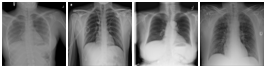}
}
\hspace{0.1in}
\subfloat{
\includegraphics[width=.15\columnwidth]{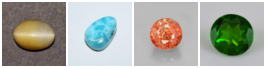}
\includegraphics[width=.15\columnwidth]{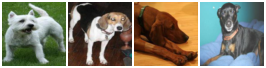}
}


\hspace{-0.1in}
\subfloat{
\includegraphics[width=.15\columnwidth]{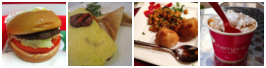}
\includegraphics[width=.15\columnwidth]{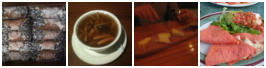}
}
\hspace{0.13in}
\subfloat{
\includegraphics[width=.15\columnwidth]{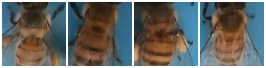}
\includegraphics[width=.15\columnwidth]{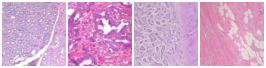}}
\hspace{0.13in}
\subfloat{
\includegraphics[width=.15\columnwidth]{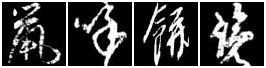}
\includegraphics[width=.15\columnwidth]{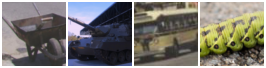}
}
\\

\hspace{-0.1in}
\subfloat{
\includegraphics[width=.15\columnwidth]{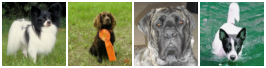}
\includegraphics[width=.15\columnwidth]{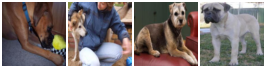}
}
\hspace{0.13in}
\subfloat{
\includegraphics[width=.15\columnwidth]{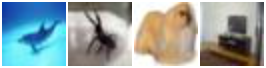}
\includegraphics[width=.15\columnwidth]{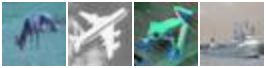}}
\hspace{0.13in}
\subfloat{
\includegraphics[width=.15\columnwidth]{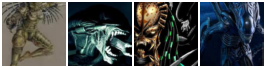}
\includegraphics[width=.15\columnwidth]{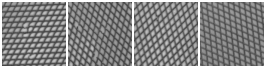}
}
\\
\vspace{-0.025in}
\caption{\small 
\textbf{Retrieved examples for meta-test datasets.} We visualize the dataset used for pretraining the retrieved model with the unseen query dataset. For more examples, see Figure~\ref{fig:unseen_retrieval_10} in Appendix.} 
\label{fig:unseen_retrieval}
\vspace{-0.15in}
\end{figure*}

\vspace{-0.05in}
\subsection{Experimental Results}

\vspace{-0.05in}
\paragraph{Meta-test Results}
\label{sec:unseen_real_world}

We compare 50-epoch accuracy between TANS and the existing NAS methods on 10 novel real-world datasets. For a fair comparison, we train PC-DARTS~\citep{xu2020pc} and DrNAS~\citep{chen2021drnas} for 10 times more epochs (500), as they only generate architectures without pretrained weights, so we train them for a sufficient amount of iterations. For FBNet and OFA (weight-sharing methods) and MetaD2A (data-driven meta-NAS), which provide the searched architectures as well as pretrained weights from ImageNet-1K, we fine-tune them on the meta-test query datasets for 50 epochs.

\begin{wrapfigure}[8]{r}{6.5cm}
    \vspace{-0.15in}
    \centering
    \includegraphics[width=1.0\linewidth]{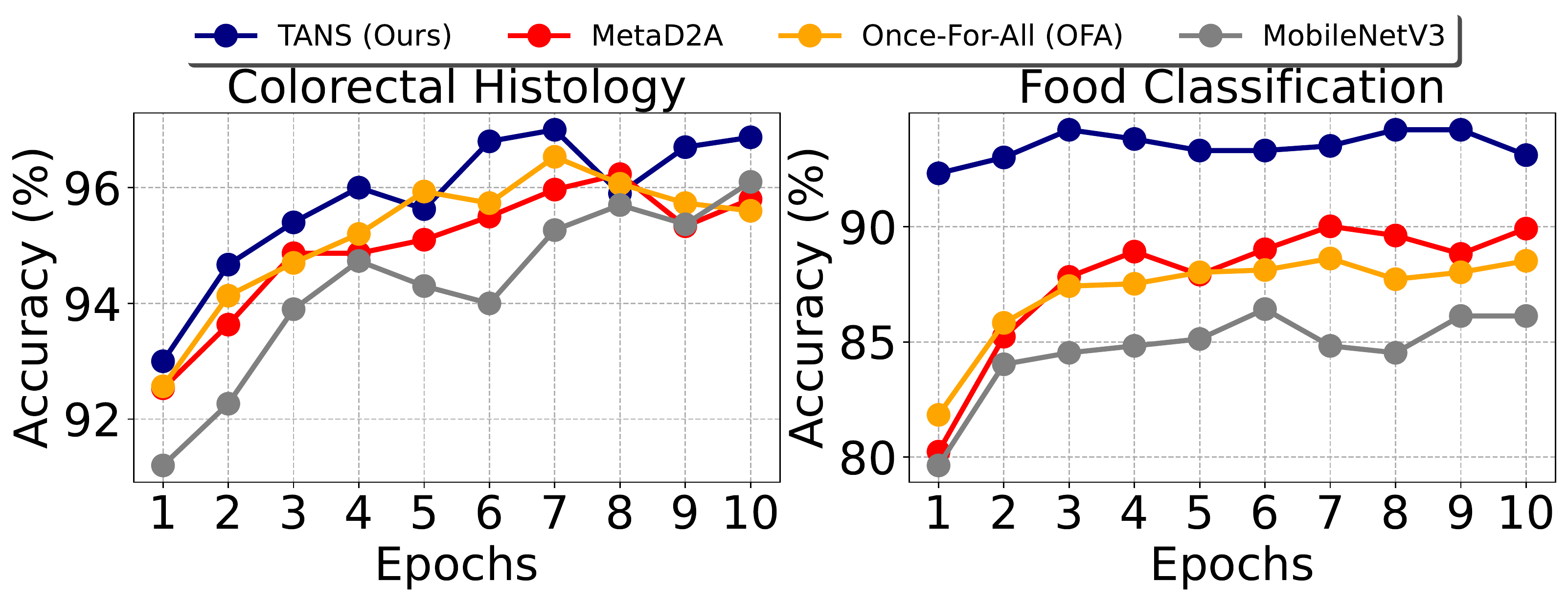}
    \vspace{-0.2in}
    \caption{\small Meta-test Accuracy Curves}
    \label{fig:meta-test-curve}
\end{wrapfigure}
As shown in Table~\ref{tbl:unseen_task}, we observe that TANS outperforms all baselines, with incomparably smaller search time and relatively smaller training time. Conventional NAS approaches such as PC-DARTS and DrNAS repeatedly require large search time for every dataset, and thus are inefficient for this practical setting with real-world datasets. FBNet, OFA, and MetaD2A are much more efficient than general NAS methods since they search for subnets within a given supernet, but obtain suboptimal performances on unseen real-world datasets as they may have largely different distributions from the dataset the supernet is trained on. In contrast, our method achieves almost zero cost in search time, and reduced training time as it fine-tunes a network pretrained on a relevant dataset. In Figure~\ref{fig:meta-test-curve}, we show the test performance curves and observe that TANS often starts with a higher starting point, and converges faster to higher accuracy. 

In Figure~\ref{fig:unseen_retrieval}, we show example images from the query and training datasets that the retrieved models are pretrained on. In most cases, our method matches semantically similar datasets to the query dataset. Even for the semantically-dissimilar cases (right column), for which our model-zoo does not contain models pretrained on datasets similar to the query, our models still outperform all other base NAS models. As such, our model effectively retrieves not only task-relevant models, but also potentially best-fitted models even trained with dissimilar datasets, for the given query datasets. We provide detailed descriptions for all query and retrieval pairs in Figure~\ref{fig:unseen_retrieval_10} of Appendix. 

We also compare with commercially available AutoML platforms, such as Microsoft Azure Custom Vision~\citep{customvision} and Google Vision Edge~\citep{visionedge}. For this experiment, we evaluate on randomly chosen five datasets (out of ten), due to excessive training costs required on AutoML platforms. As shown in Figure~\ref{fig:automl}, our method outperforms all commercial NAS/AutoML methods, with a significantly smaller total time cost. We provide more details and additional experiments, such as including real-world architectures, in Appendix~\ref{suppl:more_exps}.

\begin{figure}[t!]
\begin{tabular}{cc}
    \hspace{-0.1in}\includegraphics[width=0.49\linewidth]{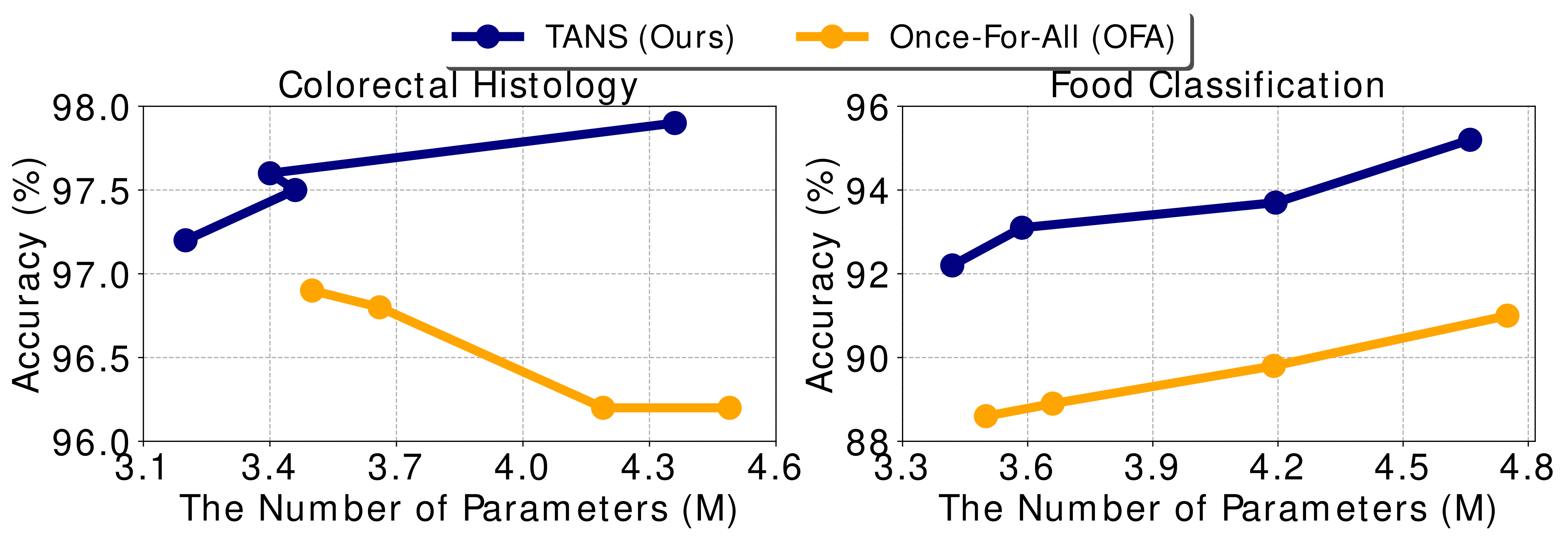}
    &  \includegraphics[width=0.49\linewidth]{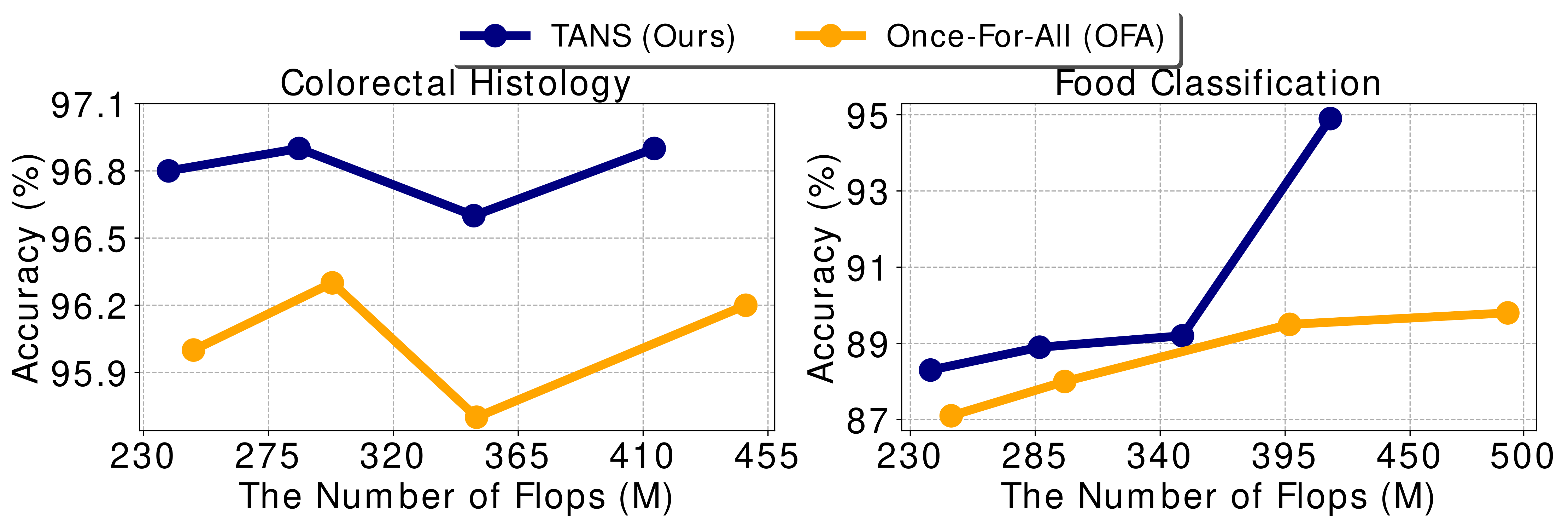}
    \\
     \small (a) Number of Parameters
    &  \small (b) FLOPs 
    \vspace{-0.05in}
\end{tabular}
    \begin{minipage}{0.375\textwidth}
        \begin{tabular}{ccc}
           \includegraphics[width=0.925\linewidth]{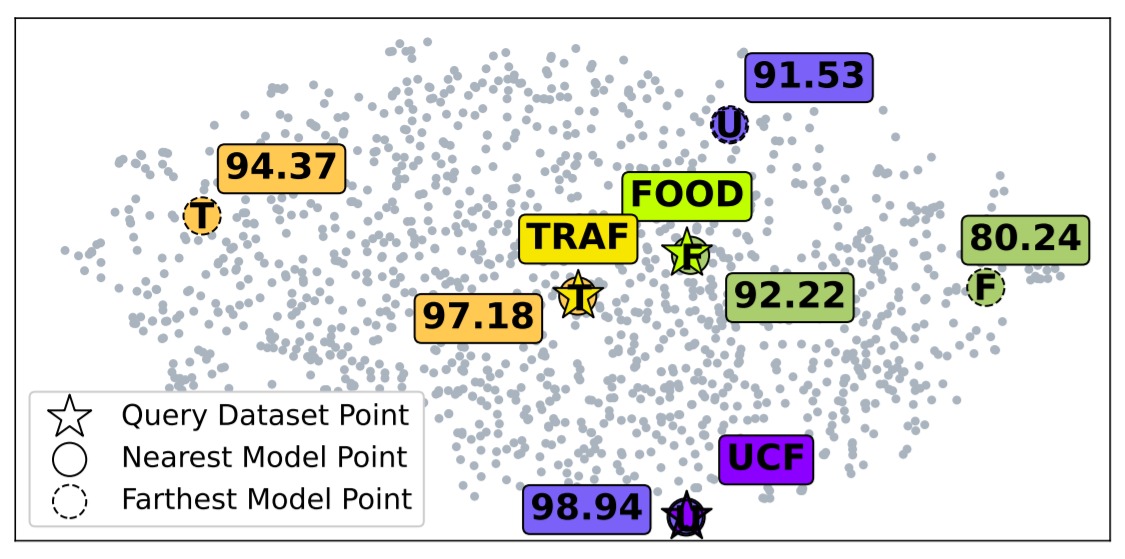}  
            \\ 
            \hspace{-0.05in} \small (c) The Cross-Modal Latent Space & 
            \hspace{-0.15in} \small (d) Effectiveness of Topology &
            \hspace{-0.05in} \small (e) Analysis on Arch. \& Parameters
        \end{tabular}
    \end{minipage}
    \begin{minipage}{0.2575\textwidth}
        \hspace{-0.05in}
        \vspace{0.1in}
        \resizebox{\textwidth}{!}{
            \begin{tabular}{l|c}
                \toprule
                \centering
                    \textbf{Method} & \textbf{Averaged Acc.}\\
                    \midrule 
                    \midrule
                    FBNet-A~\citep{wu2019fbnet}  &  93.00\small{$\pm$1.95}    \\
                    OFA~\citep{cai2020once}      &  93.89\small{$\pm$0.84}    \\
                    MetaD2A~\citep{lee2021rapid} &  95.24\small{$\pm$1.14}    \\
                    \midrule
                    TANS w/o Topol.              &  95.24\small{$\pm$0.21}    \\
                    \textbf{TANS (Ours)}         &  \textbf{96.28\small{$\pm$0.30}}    \\
                \bottomrule
            \end{tabular}
        }
    \end{minipage}\hfill
    \begin{minipage}{0.3425\textwidth}
        \vspace{-0.1in}
        \hspace{-0.05in}
        \resizebox{\textwidth}{!}{
            \begin{tabular}{l|c}
        	   \toprule
        	   \centering
        	   \textbf{Method (1/20 Model-Zoo)} & \textbf{Averaged Acc.} \\
               \midrule
               \midrule
               TANS (Ours) & 
               $95.07\%$ \\
               \midrule
               TANS w/ Random Init. &
               74.89$\%$ \\
               TANS w/ ImageNet &
               94.84$\%$ \\
               \midrule
               \textbf{TANS w/ MetaD2A Arch.} &
               \textbf{95.38$\%$} \\
        	   \bottomrule
    	   \end{tabular} 
        }
    \end{minipage}

\vspace{-0.05in}
\caption{\small \textbf{In-depth analysis of TANS:} Performance comparison with constraints, such as (a) the number of parameters and (b) FLOPs. (c) Visualization for the cross-modal latent space using T-SNE and performance comparison depending on the distance (d) Ablation study on topology information. (e) Analysis on the architecture and pretrained knowledge.}
\vspace{-0.2in}

\label{fig:convergence_and_params_acc}
\end{figure}

\vspace{-0.1in}
\paragraph{Analysis of the Architecture \& Parameters}
To see that our method is effective in retrieving networks with both optimal architecture and relevant parameters, we conduct several ablation studies. We first report the results of base models that only search for the optimal architectures. Then we provide the results of the network retrieved using a variant of our method which does not use the topology (architecture) embedding, and only uses the functional embedding $v^\tau_f$ (Tans w/o Topol). As shown in Figure~\ref{fig:convergence_and_params_acc} (d), TANS w/o Topol outperforms base NAS methods (except for MetaD2A) without considering the architectures, which shows that the performance improvement mostly comes from the knowledge transfer from the weights of the most relevant pretrained networks. However, the full TANS obtains about 1\% higher performance over TANS w/o Topol., which shows the importance of the architecture and the effectiveness of our architecture embedding. In Figure~\ref{fig:convergence_and_params_acc} (e), we further experiment with a version of our method that initializes the retrieved networks with both random weights and ImageNet pre-trained weights, using 1/20 sized model-zoo (700). We observe that they achieve lower accuracy over TANS on 10 datasets, which again shows the importance of retrieving relevant tasks' knowledge. We also construct the model-zoo by training on the architectures found by an existing NAS method (MetaD2A), and see that it further improves the performance of TANS.



\vspace{-0.1in}
\paragraph{Contraints-conditioned Retrieval}
\label{sec:params_exp}
\begin{figure}[t!]
    \begin{minipage}{0.425\textwidth}
        \begin{tabular}{cc}
            \includegraphics[width=0.95\linewidth]{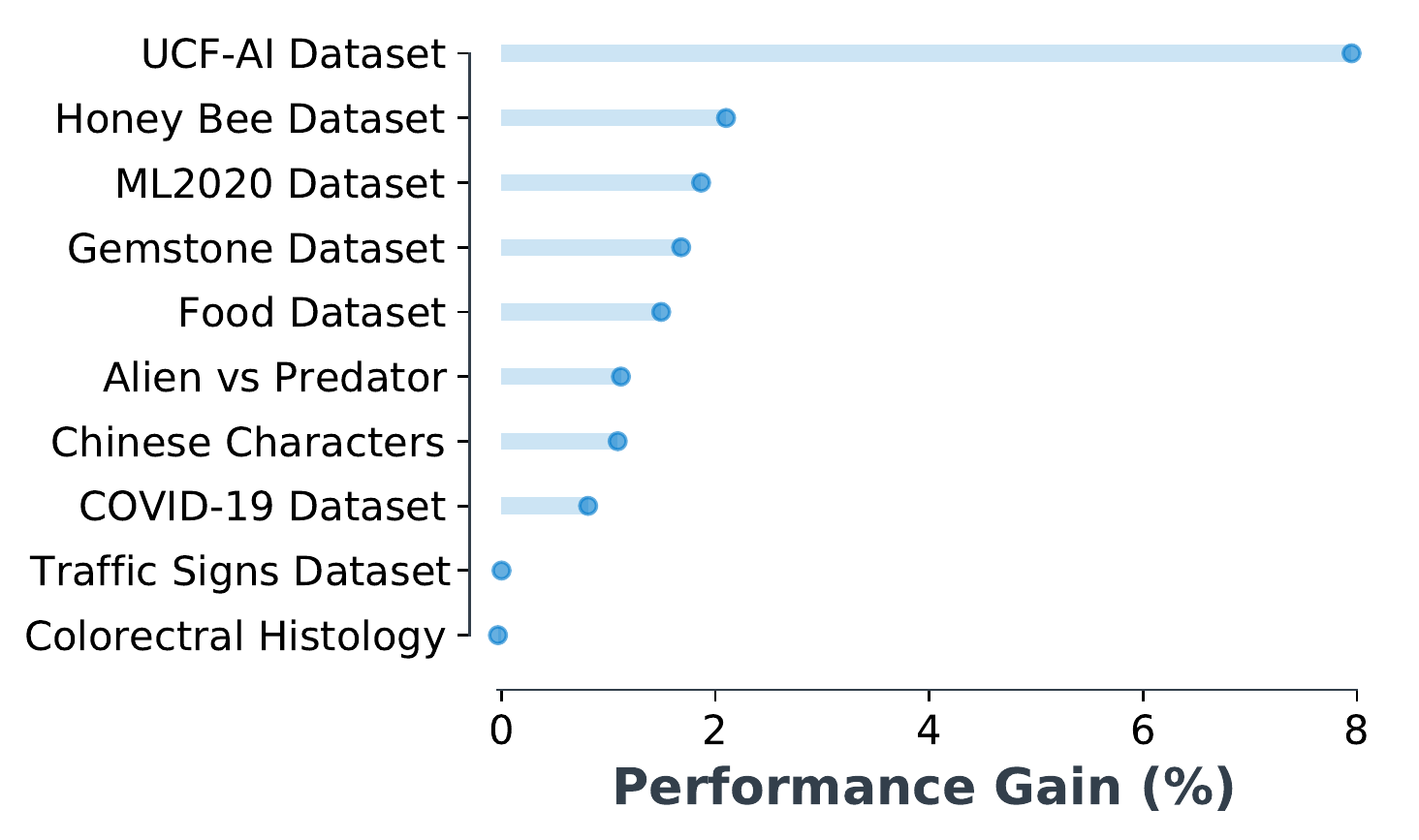}  
            \\ 
            \vspace{0.05in}
            \hspace{0.05in} \small (a) Effectiveness of Performance Predictor & 
            \hspace{0.55in}  \small (b) Analysis of Retrieval Performance 
        \end{tabular}
    \end{minipage}
    \begin{minipage}{0.575\textwidth}
        \hspace{0.05in}
        \vspace{0.2in}
        \resizebox{0.95\textwidth}{!}{
            \begin{tabular}{lccccc}
                \toprule
                    \multicolumn{1}{c}{\multirow{2}{*}{Model}} & 
                    \multicolumn{5}{c}{Performance of The Same Pair Retrieval}\\
                                      &      
                    R@1 & R@5 & R@10 & Mean & Median \\
                    \midrule 
                    \midrule
                    Random &
                    2.14           & 
                    2.86           & 
                    8.57           & 
                    69.04           & 
                    70.0          \\
                    Largest Parameter &
                    3.57           & 
                    7.14           & 
                    10.00           & 
                    51.85           & 
                    52.0           \\
                     \midrule
                     TANS + Cosine & 
                        9.29 & 12.86 &	22.14 & 46.02	& 38.0 \\ 
                    TANS + Hard Neg. & 
                       72.14 & 84.29 & 88.57 & 4.86 & 1.0 \\ 
                     \textbf{TANS + Contrastive}  &  
                     \textbf{80.71}                  & 
                     \textbf{96.43}                  & 
                     \textbf{99.29}                  & 
                     \textbf{1.90}                   & 
                     \textbf{1.0}                   \\ 
                    \midrule
                    TANS w/o Func.Emb. & 
                        5.00 &	11,43 & 18.57 & 63.20 &	63.0  \\
                    TANS w/o Predictor & 
                        80.00 &	\textbf{96.43} & 97.86 & 2.23 &	\textbf{1.0}  \\ 
                    
                 \bottomrule
                \end{tabular}
                \label{tbl:arch_meta_test}
        }
    \end{minipage}\hfill
    \\
    \vspace{0.5in}
    \begin{minipage}{0.49\textwidth}
        \resizebox{\textwidth}{!}{
            \begin{tabular}{lccccc}
                \toprule
                    \multicolumn{1}{l}{\multirow{2}{*}{Method}} & 
                    \multicolumn{5}{c}{MSE on 5 Meta-test Datasets}
                    \\
                    
                    & Food & Gemstones & Dogs & A. vs P. & COVID-19 \\
                    \midrule 
                    \midrule
                    
                    Predictor w/o $v_q^{\tau}$ &
                    0.0178 & 0.0782 & 0.0194 & 0.0185 & 0.0418 \\
                    
                    Predictor w/o $v_f^{\tau}$ &
                    0.0188 & \textbf{0.0323} & 0.0016 & 0.0652 & 0.0328 \\
                    
                    \midrule
                    \textbf{Predictor (Ours)}&
                    \textbf{0.0036} & 0.0338 & \textbf{0.0013} & 
                    \textbf{0.0028} & \textbf{0.0233} \\ 
                    
                 \bottomrule
                \end{tabular}
                \label{tbl:arch_meta_test}
        }
    \end{minipage}\hfill
    \begin{minipage}{0.49\textwidth}
        \resizebox{\textwidth}{!}{
            \begin{tabular}{lccccc}
                \toprule
                    \multicolumn{1}{l}{\multirow{2}{*}{Method}} & 
                    \multicolumn{5}{c}{MSE on 5 Meta-test Datasets}
                    \\
                    
                    & Food & Gemstones & Dogs & A. vs P. & COVID-19 \\
                    \midrule 
                    \midrule
                    
                    Random Estimation &
                    0.1619 & 0.1081 & 0.1348 & 0.2609 & 0.2928 \\
                    
                    1/100 Model-Zoo &
                    0.0088 & 0.0369 & 0.0034 & 0.0077 & 0.0241 \\
                    
                    \midrule
                    \textbf{Top 10 Retrievals}&
                    \textbf{0.0036} & \textbf{0.0338} & \textbf{0.0013} & \textbf{0.0028} & \textbf{0.0233} \\ 
                    
                 \bottomrule
                \end{tabular}
                \label{tbl:arch_meta_test}
        }
    \end{minipage}\hfill
    \vspace{-.45in}
    \begin{tabular}{cc}
        \hspace{0.3in} \small (c) Analysis of Performance Predictor & 
        \hspace{0.45in} \small (d) Effectiveness of Performance Predictor
    \end{tabular}
    \vspace{-0.05in}
    \caption{\small \textbf{In-depth analysis of TANS (2):} (a) The effectiveness of meta-performance predictor. (b) Ablation study on retrieval performance. Additional (c) analysis and (d) the effectiveness of our performance predictor.}
    \vspace{-0.2in}
    
    \label{fig:indepth-analysis-2}
\end{figure}

TANS can retrieve models with a given dataset and additional constraints, such as the number of the parameters or the computations (in FLOP). This is practically important since we may need a network with less memory and computation overhead depending on the hardware device. This can be done by filtering networks that satisfy the given conditions among the candidate networks retrieved. For this experiment, we compare against OFA that performs the same constrained search, as other baselines do not straightforwardly handle this scenario. As shown in Figure~\ref{fig:convergence_and_params_acc} (a) and (b), we observe that the network retrieved with TANS consistently outperforms the network searched with OFA under varying parameters and computations constraints. Such constrained search is straightforward with our method since our retrieval-based method allows us to search from the database consisting of networks with varying architectures and sizes.

\vspace{-0.1in}
\paragraph{Analysis of the Cross-Modal Retrieval Space} We further examine the learned cross-modal space. We first visualize the meta-learned latent space in Figure~\ref{fig:convergence_and_params_acc} (c) with randomly sampled 1,400 models among 14,000 models in the model-zoo. We observe that the network whose embeddings are the closest to the query dataset achieves higher performance on it, compared to networks embedded the farthest from it. For example, the accuracy of the closet network point for UCF-AI is $98.94\%$ while the farthest network only achieves $91.53\%$.

\begin{wraptable}[5]{r}{6.5cm}
    \vspace{-0.25in}
    \centering
    \small 
    \caption{Latent Distance and Performance}
    \vspace{-0.05in}
    \resizebox{0.475\columnwidth}{!}{
        \vspace{-0.2in}
            \begin{tabular}{ccccc}
                \toprule
                    \multicolumn{5}{c}{Spearman's Correlation on 5 Meta-test Datasets} \\
                    Food & Drawing & Chinese & A. vs P. & Colorectal \\
                    \midrule 
                    \midrule
                    
                    0.752           & 
                    0.583           & 
                    0.322          & 
                    0.214           & 
                    0.213          \\
                    
                 \bottomrule
                \end{tabular}
                \label{tbl:corr}
        }
    \vspace{-0.1in}
\end{wraptable}

We also show Spearman correlation scores on 5 meta-test datasets in Table~\ref{tbl:corr}. Measuring correlation with the distances is not directly compatible with our contrastive loss, since the negative examples (models that achieve low performance on the target dataset) are pushed away from the query point, without a meaningful ranking between the negative instances. To obtain a latent space where the negative examples are also well-ranked, we should replace the contrastive loss with a ranking loss instead, but this will not be very meaningful. Hence, to effectively validate our latent space with correlation metric, we rather select clusters, such that 50 models around the query point and another 50 models around the farthest points, thus using a total of 100 models to report the correlation scores. In the Table~\ref{tbl:corr}, we show the correlation scores of these 100 models on the five unseen datasets. For Food dataset (reported as "hard" dataset in Table~\ref{tbl:unseen_task}), the correlation scores are shown to be high. On the other hand, for Colorectal Histology dataset (reported as "easy" dataset), the correlation scores are low as any model can obtain good performance on it, which makes the performance gap across models small. In sum, as the task (dataset) becomes more difficult, we can observe a higher correlation in the latent space between the distance and the rank. 

\vspace{-0.05in}

\paragraph{Meta-performance Predictor} The role of the proposed performance predictor is not only guiding the model and query encoder to learn the better latent space but also selecting the best model among retrieval candidates. To verify its effectiveness, we measure the performance gap between the top-1 retrieved model w/o the predictor and the model with the highest scores selected using the predictor among retrieved candidates. As shown in Figure~\ref{fig:indepth-analysis-2} (a), there are $1.5\%p$ - $8\%p$ performance gains on the meta-test datasets. The top-1 retrieved model from the model zoo with TANS may not be always optimal for an unseen dataset, and our performance predictor remedies this issue by selecting the best-fitted model based on its estimation. We also examine ablation study for our performance predictor. Please note that we do not use ranking loss which does not rank the negative examples. Thus we use Mean Squared Error (MSE) scores. We retrieve the top 10 most relevant models for an unseen query datasets and then compute the MSE between the estimated scores and the actual ground truth accuracies. As shown in Figure~\ref{fig:indepth-analysis-2} (c), we observe that removing either query or model embeddings degrades performance compared to the predictor taking both embeddings. It is natural that, with only model or query information, it is difficult to correctly estimate the accuracy since the predictor fails to recognize what or where to learn. Also, we report the MSE between the predicted performance using the predictor and the ground truth performance of each model for the entire set of pretrained models from a smaller model zoo in Figure~\ref{fig:indepth-analysis-2} (d). Although the performance predictor achieves slightly higher MSE scores for this experiment compared to the MSE obtained on the top-10 retrieved models (which are the most important), the MSE scores are still meaningfully low, which implies that our performance model successfully works even 
toward the entire model-zoo.

\vspace{-0.1in}
\paragraph{Retrieval Performance}
\label{sec:model_retrieval}
We also verify whether our model successfully retrieves the same paired models when the correspondent \textbf{meta-train} datasets are given (we use unseen instances that are \textbf{not used} when training the encoders.) For the evaluation metric, we use recall at \textit{k} (\textbf{R@\textit{k}}) which indicates the percentage of the correct models retrieved among the top-\textit{k} candidates for the unseen query instances, where \textit{k} is set to $1$, $5$, and $10$. Also, we report the mean and median ranking of the correct network among all networks for the unseen query. In Figure~\ref{fig:indepth-analysis-2} (b), the largest parameter selection strategy shows poor performances on the retrieval problem, suggesting that simply selecting the largest network is not a suitable choice for real-world tasks. In addition, compared to cosine similarity learning, the proposed meta-contrastive learning allows the model to learn significantly improved discriminative latent space for cross-modal retrieval. Moreover, without our performance predictor, we observe that TANS achieves slightly lower performance, while it is significantly degenerated when training without functional embeddings.

\vspace{-0.1in}
\paragraph{Analysis of Model-Zoo Construction} 
\begin{wrapfigure}[13]{r}{5cm}
    \vspace{-0.2in}
    \small
    \centering
    \resizebox{0.36\textwidth}{!}{
        \hspace{-0.125in}
        \begin{tabular}{l|c|c}
        	   \toprule 
        	   \centering
        	   \multirow{2}{*}{Method}                 & 
        	   Architecture                            &
        	   Retrieved Dataset                       \\
        	   &
        	   Search Cost                             &
        	   Pre-training Cost                       \\
              \midrule
              \midrule
              Conventional NAS                         & 
              O(N)                                     &
              O(N)                                     \\
              MetaD2A                                  & 
              O(1)                                     &
              O(N)                                     \\
              \textbf{TANS (Ours)}                     &
              \textbf{O(1)}                            &
              \textbf{O(1)}                            \\
        	  \bottomrule
        \end{tabular} 
    }
    \includegraphics[width=0.34\textwidth]{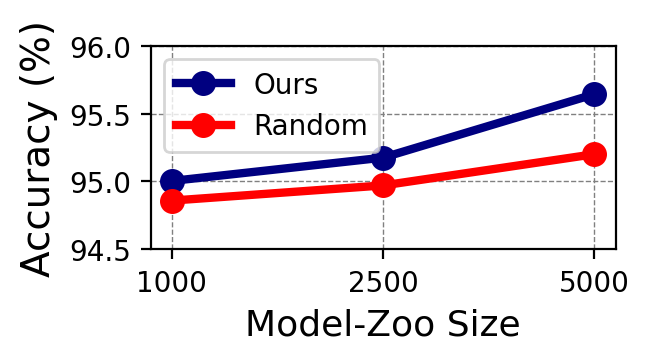}
    \vspace{-0.1in}
    \caption{\small Search \& Pretraining Costs (Top), Model-Zoo Analysis (Bottom)}
    \label{fig:model_zoo}
\end{wrapfigure}

Unlike most existing NAS methods, which repeatedly search the optimal architectures \textit{per dataset} from their search spaces, TANS do not need to perform such repetitive search procedure, once the model-zoo is built beforehand. We are able to adaptively retrieve the relevant pretrained models for \textbf{any number of datasets} from our model-zoo, with almost zero search costs. Formally, TANS reduces the time complexities of both search cost and pre-training cost {\textbf{from O(N) to O(1)}}, where $N$ is the number of datasets, as shown in Figure~\ref{fig:model_zoo} (Top). Furthermore, a model zoo constructed using our efficient construction algorithm, introdcued in Section~\ref{sec:modelzoocons}, yields models with higher performance on average, compared to the random sampling strategy when the size of the model-zoo is the same as shown in Figure~\ref{fig:model_zoo} (Bottom). 




\vspace{-0.1in}
\section{Conclusion}
\label{conclusion}
\vspace{-0.05in}
We propose a novel Task-Adaptive Neural Network Search framework (TANS), that instantly retrieves a relevant pre-trained network for an unseen dataset, based on the cross-modal retrieval of dataset-network pairs. We train this retrieval model via amortized meta-learning of the cross-modal latent space with contrastive learning, to maximize the similarity between the positive dataset-network pairs, and minimize the similarity between the negative pairs. We train our framework on a model zoo consisting of diverse pretrained networks, and validate its performance on ten unseen datastes. The results show that the proposed TANS rapidly searches and trains a well-fitted network for unseen datasets with almost no architecture search cost and significantly fewer fine-tuning steps to reach the target performance, compared to other NAS methods. We discuss the \textbf{limitation and the societal impact} of our work in 
Appendix~\ref{sec:discuss}.


\clearpage

\small 
\paragraph{Acknowledgement} This work was supported by AITRICS and by Institute of Information \& communications Technology Planning \& Evaluation (IITP) grant funded by the Korea government(MSIT) (No.2019-0-00075, Artificial Intelligence Graduate School Program(KAIST))

\bibliography{reference}

\begin{thebibliography}{77}
\providecommand{\natexlab}[1]{#1}
\providecommand{\url}[1]{\texttt{#1}}
\expandafter\ifx\csname urlstyle\endcsname\relax
  \providecommand{\doi}[1]{doi: #1}\else
  \providecommand{\doi}{doi: \begingroup \urlstyle{rm}\Url}\fi

\bibitem[cus()]{customvision}
Microsoft azure custom vision.
\newblock \url{https://www.customvision.ai/}.

\bibitem[vis()]{visionedge}
Google automl vision edge.
\newblock \url{https://cloud.google.com/vision/automl}.

\bibitem[Agarwal et~al.(2018)Agarwal, Beygelzimer, Dudík, Langford, and
  Wallach]{agarwal2018reductions}
Alekh Agarwal, Alina Beygelzimer, Miroslav Dudík, John Langford, and Hanna
  Wallach.
\newblock A reductions approach to fair classification, 2018.

\bibitem[Bello(2021)]{bello2021lambdanetworks}
Irwan Bello.
\newblock Lambdanetworks: Modeling long-range interactions without attention.
\newblock In \emph{International Conference on Learning Representations}, 2021.
\newblock URL \url{https://openreview.net/forum?id=xTJEN-ggl1b}.

\bibitem[Biscani and Izzo(2020)]{Biscani2020}
Francesco Biscani and Dario Izzo.
\newblock A parallel global multiobjective framework for optimization: pagmo.
\newblock \emph{Journal of Open Source Software}, 2020.

\bibitem[Brock et~al.(2018)Brock, Lim, Ritchie, and Weston]{brock2018smash}
Andrew Brock, Theodore Lim, James~M. Ritchie, and Nick Weston.
\newblock {SMASH:} one-shot model architecture search through hypernetworks.
\newblock In \emph{6th International Conference on Learning Representations,
  {ICLR} 2018, Vancouver, BC, Canada, April 30 - May 3, 2018, Conference Track
  Proceedings}, 2018.

\bibitem[Brock et~al.(2021)Brock, De, Smith, and Simonyan]{brock2021high}
Andrew Brock, Soham De, Samuel~L Smith, and Karen Simonyan.
\newblock High-performance large-scale image recognition without normalization.
\newblock \emph{arXiv preprint arXiv:2102.06171}, 2021.

\bibitem[Cai et~al.(2020)Cai, Gan, Wang, Zhang, and Han]{cai2020once}
Han Cai, Chuang Gan, Tianzhe Wang, Zhekai Zhang, and Song Han.
\newblock Once for all: Train one network and specialize it for efficient
  deployment.
\newblock In \emph{International Conference on Learning Representations}, 2020.

\bibitem[Chang et~al.(2020)Chang, Yu, Chang, Yang, and Kumar]{chang2020ir1}
Wei{-}Cheng Chang, Felix~X. Yu, Yin{-}Wen Chang, Yiming Yang, and Sanjiv Kumar.
\newblock Pre-training tasks for embedding-based large-scale retrieval.
\newblock In \emph{8th International Conference on Learning Representations,
  {ICLR} 2020, Addis Ababa, Ethiopia, April 26-30, 2020}, 2020.

\bibitem[Chen et~al.(2021)Chen, Wang, Cheng, Tang, and Hsieh]{chen2021drnas}
Xiangning Chen, Ruochen Wang, Minhao Cheng, Xiaocheng Tang, and Cho-Jui Hsieh.
\newblock Dr{\{}nas{\}}: Dirichlet neural architecture search.
\newblock In \emph{International Conference on Learning Representations}, 2021.

\bibitem[Chen et~al.(2019)Chen, Meng, Zhang, Xiang, Huang, Mu, and
  Wang]{chen2019evolve}
Yukang Chen, Gaofeng Meng, Qian Zhang, Shiming Xiang, Chang Huang, Lisen Mu,
  and Xinggang Wang.
\newblock {RENAS:} reinforced evolutionary neural architecture search.
\newblock In \emph{{IEEE} Conference on Computer Vision and Pattern
  Recognition, {CVPR} 2019, Long Beach, CA, USA, June 16-20, 2019}, pages
  4787--4796. Computer Vision Foundation / {IEEE}, 2019.

\bibitem[Cotter et~al.(2018)Cotter, Gupta, Jiang, Srebro, Sridharan, Wang,
  Woodworth, and You]{cotter2018training}
Andrew Cotter, Maya Gupta, Heinrich Jiang, Nathan Srebro, Karthik Sridharan,
  Serena Wang, Blake Woodworth, and Seungil You.
\newblock Training well-generalizing classifiers for fairness metrics and other
  data-dependent constraints, 2018.

\bibitem[Deng et~al.(2009)Deng, Dong, Socher, Li, Li, and
  Fei-Fei]{deng2009imagenet}
Jia Deng, Wei Dong, Richard Socher, Li-Jia Li, Kai Li, and Li~Fei-Fei.
\newblock Imagenet: A large-scale hierarchical image database.
\newblock In \emph{2009 IEEE conference on computer vision and pattern
  recognition}, pages 248--255. Ieee, 2009.

\bibitem[Devaraj et~al.(2020)Devaraj, Murugaboopathi, Elhoseny, Shankar, Min,
  Moon, and Joshi]{devaraj2020efficient}
A~Francis~Saviour Devaraj, G~Murugaboopathi, Mohamed Elhoseny, K~Shankar,
  Kyungbok Min, Hyeonjoon Moon, and Gyanendra~Prasad Joshi.
\newblock An efficient framework for secure image archival and retrieval system
  using multiple secret share creation scheme.
\newblock \emph{IEEE Access}, 8:\penalty0 144310--144320, 2020.

\bibitem[Dosovitskiy et~al.(2021)Dosovitskiy, Beyer, Kolesnikov, Weissenborn,
  Zhai, Unterthiner, Dehghani, Minderer, Heigold, Gelly, Uszkoreit, and
  Houlsby]{dosovitskiy2021an}
Alexey Dosovitskiy, Lucas Beyer, Alexander Kolesnikov, Dirk Weissenborn,
  Xiaohua Zhai, Thomas Unterthiner, Mostafa Dehghani, Matthias Minderer, Georg
  Heigold, Sylvain Gelly, Jakob Uszkoreit, and Neil Houlsby.
\newblock An image is worth 16x16 words: Transformers for image recognition at
  scale.
\newblock In \emph{International Conference on Learning Representations}, 2021.
\newblock URL \url{https://openreview.net/forum?id=YicbFdNTTy}.

\bibitem[Dwork et~al.(2012)Dwork, Hardt, Pitassi, Reingold, and
  Zemel]{10.1145/2090236.2090255}
Cynthia Dwork, Moritz Hardt, Toniann Pitassi, Omer Reingold, and Richard Zemel.
\newblock Fairness through awareness.
\newblock In \emph{Proceedings of the 3rd Innovations in Theoretical Computer
  Science Conference}, ITCS '12, page 214–226, New York, NY, USA, 2012.
  Association for Computing Machinery.
\newblock ISBN 9781450311151.
\newblock \doi{10.1145/2090236.2090255}.
\newblock URL \url{https://doi.org/10.1145/2090236.2090255}.

\bibitem[Elsken et~al.(2020)Elsken, Staffler, Metzen, and
  Hutter]{elsken2020metanas3}
Thomas Elsken, Benedikt Staffler, Jan~Hendrik Metzen, and Frank Hutter.
\newblock Meta-learning of neural architectures for few-shot learning.
\newblock In \emph{2020 {IEEE/CVF} Conference on Computer Vision and Pattern
  Recognition, {CVPR} 2020, Seattle, WA, USA, June 13-19, 2020}, pages
  12362--12372. {IEEE}, 2020.

\bibitem[Engilberge et~al.(2018)Engilberge, Chevallier, P{\'{e}}rez, and
  Cord]{Martin18}
Martin Engilberge, Louis Chevallier, Patrick P{\'{e}}rez, and Matthieu Cord.
\newblock Finding beans in burgers: Deep semantic-visual embedding with
  localization.
\newblock In \emph{The IEEE Conference on Computer Vision and Pattern
  Recognition (CVPR)}, 2018.

\bibitem[Faghri et~al.(2018)Faghri, Fleet, Kiros, and Fidler]{Faghri18}
Fartash Faghri, David~J. Fleet, Jamie Kiros, and Sanja Fidler.
\newblock {VSE++:} improving visual-semantic embeddings with hard negatives.
\newblock In \emph{British Machine Vision Conference 2018, {BMVC}}, 2018.

\bibitem[Finn et~al.(2017)Finn, Abbeel, and Levine]{finn2017model}
Chelsea Finn, Pieter Abbeel, and Sergey Levine.
\newblock Model-agnostic meta-learning for fast adaptation of deep networks.
\newblock In \emph{International Conference on Machine Learning (ICML)}, 2017.

\bibitem[Gordo et~al.(2017)Gordo, Almaz\'{a}n, Revaud, and
  Larlus]{gordo2017end}
Albert Gordo, Jon Almaz\'{a}n, Jerome Revaud, and Diane Larlus.
\newblock End-to-end learning of deep visual representations for image
  retrieval.
\newblock \emph{International Journal of Computer Vision}, 124\penalty0
  (2):\penalty0 237--254, September 2017.

\bibitem[Hardt et~al.(2016)Hardt, Price, and Srebro]{Hardt2016EqualityOO}
Moritz Hardt, Eric Price, and Nathan Srebro.
\newblock Equality of opportunity in supervised learning.
\newblock In \emph{NIPS}, 2016.

\bibitem[He et~al.(2016)He, Zhang, Ren, and Sun]{he2016resnet}
Kaiming He, Xiangyu Zhang, Shaoqing Ren, and Jian Sun.
\newblock Deep residual learning for image recognition.
\newblock In \emph{2016 {IEEE} Conference on Computer Vision and Pattern
  Recognition, {CVPR}}, 2016.

\bibitem[Hornik(1991)]{hornik91universal}
Kurt Hornik.
\newblock Approximation capabilities of multilayer feedforward networks.
\newblock \emph{Neural Networks}, 4\penalty0 (2):\penalty0 251--257, 1991.

\bibitem[Hornik et~al.(1989)Hornik, Stinchcombe, and White]{Hornik89universal}
Kurt Hornik, Maxwell~B. Stinchcombe, and Halbert White.
\newblock Multilayer feedforward networks are universal approximators.
\newblock \emph{Neural Networks}, 2\penalty0 (5):\penalty0 359--366, 1989.

\bibitem[Howard et~al.(2019)Howard, Sandler, Chu, Chen, Chen, Tan, Wang, Zhu,
  Pang, Vasudevan, et~al.]{howard2019searching}
Andrew Howard, Mark Sandler, Grace Chu, Liang-Chieh Chen, Bo~Chen, Mingxing
  Tan, Weijun Wang, Yukun Zhu, Ruoming Pang, Vijay Vasudevan, et~al.
\newblock Searching for mobilenetv3.
\newblock In \emph{Proceedings of the IEEE International Conference on Computer
  Vision}, pages 1314--1324, 2019.

\bibitem[Howard et~al.(2017)Howard, Zhu, Chen, Kalenichenko, Wang, Weyand,
  Andreetto, and Adam]{howard2017mobilenet}
Andrew~G. Howard, Menglong Zhu, Bo~Chen, Dmitry Kalenichenko, Weijun Wang,
  Tobias Weyand, Marco Andreetto, and Hartwig Adam.
\newblock Mobilenets: Efficient convolutional neural networks for mobile vision
  applications.
\newblock \emph{arXiv preprint arXiv:1704.04861}, 2017.

\bibitem[Iandola et~al.(2016)Iandola, Han, Moskewicz, Ashraf, Dally, and
  Keutzer]{iandola2016squeezenet}
Forrest~N Iandola, Song Han, Matthew~W Moskewicz, Khalid Ashraf, William~J
  Dally, and Kurt Keutzer.
\newblock Squeezenet: Alexnet-level accuracy with 50x fewer parameters and< 0.5
  mb model size.
\newblock \emph{arXiv preprint arXiv:1602.07360}, 2016.

\bibitem[Krizhevsky et~al.(2012)Krizhevsky, Sutskever, and
  Hinton]{krizhevsky2012imagenet}
Alex Krizhevsky, Ilya Sutskever, and Geoffrey~E Hinton.
\newblock Imagenet classification with deep convolutional neural networks.
\newblock \emph{Advances in neural information processing systems},
  25:\penalty0 1097--1105, 2012.

\bibitem[Lee et~al.(2020)Lee, Lee, Na, Kim, Park, Yang, and Hwang]{lee2020l2b}
Hae~Beom Lee, Hayeon Lee, Donghyun Na, Saehoon Kim, Minseop Park, Eunho Yang,
  and Sung~Ju Hwang.
\newblock Learning to balance: Bayesian meta-learning for imbalanced and
  out-of-distribution tasks.
\newblock In \emph{International Conference on Learning Representations
  (ICLR)}, 2020.

\bibitem[Lee et~al.(2021)Lee, Hyung, and Hwang]{lee2021rapid}
Hayeon Lee, Eunyoung Hyung, and Sung~Ju Hwang.
\newblock Rapid neural architecture search by learning to generate graphs from
  datasets.
\newblock In \emph{International Conference on Learning Representations}, 2021.

\bibitem[Lee et~al.(2018)Lee, Chen, Hua, Hu, and He]{lee2018crossmodal2}
Kuang{-}Huei Lee, Xi~Chen, Gang Hua, Houdong Hu, and Xiaodong He.
\newblock Stacked cross attention for image-text matching.
\newblock In \emph{Computer Vision - {ECCV} 2018 - 15th European Conference,
  Munich, Germany, September 8-14, 2018, Proceedings, Part {IV}}, volume 11208
  of \emph{Lecture Notes in Computer Science}, pages 212--228. Springer, 2018.

\bibitem[Lee et~al.(2019)Lee, Maji, Ravichandran, and Soatto]{lee2019meta}
Kwonjoon Lee, Subhransu Maji, Avinash Ravichandran, and Stefano Soatto.
\newblock Meta-learning with differentiable convex optimization.
\newblock In \emph{Proceedings of the IEEE Conference on Computer Vision and
  Pattern Recognition (CVPR)}, 2019.

\bibitem[Lian et~al.(2020)Lian, Zheng, Xu, Lu, Lin, Zhao, Huang, and
  Gao]{lian2020metanas2}
Dongze Lian, Yin Zheng, Yintao Xu, Yanxiong Lu, Leyu Lin, Peilin Zhao, Junzhou
  Huang, and Shenghua Gao.
\newblock Towards fast adaptation of neural architectures with meta learning.
\newblock In \emph{8th International Conference on Learning Representations,
  {ICLR} 2020, Addis Ababa, Ethiopia, April 26-30, 2020}, 2020.

\bibitem[Liu et~al.(2019)Liu, Simonyan, and Yang]{liu2018darts}
Hanxiao Liu, Karen Simonyan, and Yiming Yang.
\newblock Darts: Differentiable architecture search.
\newblock In \emph{In International Conference on Learning Representations
  (ICLR)}, 2019.

\bibitem[Louizos et~al.(2017)Louizos, Swersky, Li, Welling, and
  Zemel]{louizos2017variational}
Christos Louizos, Kevin Swersky, Yujia Li, Max Welling, and Richard Zemel.
\newblock The variational fair autoencoder, 2017.

\bibitem[Lu et~al.(2020)Lu, Deb, Goodman, Banzhaf, and
  Boddeti]{lu2020nsganetv2}
Zhichao Lu, Kalyanmoy Deb, Erik Goodman, Wolfgang Banzhaf, and Vishnu~Naresh
  Boddeti.
\newblock Nsganetv2: Evolutionary multi-objective surrogate-assisted neural
  architecture search.
\newblock In \emph{European Conference on Computer Vision}, pages 35--51.
  Springer, 2020.

\bibitem[Luo et~al.(2018)Luo, Tian, Qin, Chen, and Liu]{luo2018neural}
Renqian Luo, Fei Tian, Tao Qin, Enhong Chen, and Tie-Yan Liu.
\newblock Neural architecture optimization.
\newblock In \emph{Advances in neural information processing systems
  (NeurIPS)}, 2018.

\bibitem[Ma et~al.(2018)Ma, Zhang, Zheng, and Sun]{ma2018shufflenet}
Ningning Ma, Xiangyu Zhang, Hai-Tao Zheng, and Jian Sun.
\newblock Shufflenet v2: Practical guidelines for efficient cnn architecture
  design.
\newblock In \emph{Proceedings of the European conference on computer vision
  (ECCV)}, pages 116--131, 2018.

\bibitem[Nichol et~al.(2018)Nichol, Achiam, and Schulman]{nichol2018first}
Alex Nichol, Joshua Achiam, and John Schulman.
\newblock On first-order meta-learning algorithms.
\newblock \emph{arXiv preprint arXiv:1803.02999}, 2018.

\bibitem[Nowak et~al.(2014)Nowak, Märtens, and Izzo]{nowak_empirical_2014}
Krzysztof Nowak, Marcus Märtens, and Dario Izzo.
\newblock Empirical {Performance} of the {Approximation} of the {Least}
  {Hypervolume} {Contributor}.
\newblock In Thomas Bartz-Beielstein, Jürgen Branke, Bogdan Filipič, and Jim
  Smith, editors, \emph{Parallel {Problem} {Solving} from {Nature} – {PPSN}
  {XIII}}, Lecture {Notes} in {Computer} {Science}, pages 662--671, Cham, 2014.
  Springer International Publishing.
\newblock ISBN 978-3-319-10762-2.
\newblock \doi{10.1007/978-3-319-10762-2_65}.

\bibitem[Pham et~al.(2018)Pham, Guan, Zoph, Le, and Dean]{pham2918enas}
Hieu Pham, Melody~Y. Guan, Barret Zoph, Quoc~V. Le, and Jeff Dean.
\newblock Efficient neural architecture search via parameter sharing.
\newblock In \emph{Proceedings of the 35th International Conference on Machine
  Learning, {ICML} 2018, Stockholmsm{\"{a}}ssan, Stockholm, Sweden, July 10-15,
  2018}, volume~80 of \emph{Proceedings of Machine Learning Research}, pages
  4092--4101. {PMLR}, 2018.

\bibitem[Real et~al.(2019)Real, Aggarwal, Huang, and Le]{real2019regularized}
Esteban Real, Alok Aggarwal, Yanping Huang, and Quoc~V Le.
\newblock Regularized evolution for image classifier architecture search.
\newblock In \emph{Proceedings of the aaai conference on artificial
  intelligence}, volume~33, pages 4780--4789, 2019.

\bibitem[Ren et~al.(2020)Ren, Xiao, Chang, Huang, Li, Chen, and
  Wang]{ren2020survey}
Pengzhen Ren, Yun Xiao, Xiaojun Chang, Po{-}Yao Huang, Zhihui Li, Xiaojiang
  Chen, and Xin Wang.
\newblock A comprehensive survey of neural architecture search: Challenges and
  solutions, 2020.

\bibitem[Sandler et~al.(2018)Sandler, Howard, Zhu, Zhmoginov, and
  Chen]{sandler2018mobilenetv2}
Mark Sandler, Andrew Howard, Menglong Zhu, Andrey Zhmoginov, and Liang-Chieh
  Chen.
\newblock Mobilenetv2: Inverted residuals and linear bottlenecks.
\newblock In \emph{Proceedings of the IEEE conference on computer vision and
  pattern recognition}, pages 4510--4520, 2018.

\bibitem[Shaw et~al.(2019)Shaw, Wei, Liu, Song, and Dai]{shaw2019metanas1}
Albert Shaw, Wei Wei, Weiyang Liu, Le~Song, and Bo~Dai.
\newblock Meta architecture search.
\newblock In \emph{Advances in Neural Information Processing Systems 32: Annual
  Conference on Neural Information Processing Systems 2019, NeurIPS 2019,
  December 8-14, 2019, Vancouver, BC, Canada}, pages 11225--11235, 2019.

\bibitem[Simonyan and Zisserman(2015)]{simonyan2014vgg}
Karen Simonyan and Andrew Zisserman.
\newblock Very deep convolutional networks for large-scale image recognition.
\newblock In \emph{3rd International Conference on Learning Representations,
  {ICLR} 2015, San Diego, CA, USA, May 7-9, 2015, Conference Track
  Proceedings}, 2015.

\bibitem[Snell et~al.(2017)Snell, Swersky, and Zemel]{snell2017prototypical}
Jake Snell, Kevin Swersky, and Richard Zemel.
\newblock Prototypical networks for few-shot learning.
\newblock In \emph{Advances in neural information processing systems (NIPS)},
  2017.

\bibitem[Srinivas et~al.(2021)Srinivas, Lin, Parmar, Shlens, Abbeel, and
  Vaswani]{srinivas2021bottleneck}
Aravind Srinivas, Tsung-Yi Lin, Niki Parmar, Jonathon Shlens, Pieter Abbeel,
  and Ashish Vaswani.
\newblock Bottleneck transformers for visual recognition.
\newblock \emph{arXiv preprint arXiv:2101.11605}, 2021.

\bibitem[Szegedy et~al.(2015)Szegedy, Liu, Jia, Sermanet, Reed, Anguelov,
  Erhan, Vanhoucke, and Rabinovich]{szegedy2015going}
Christian Szegedy, Wei Liu, Yangqing Jia, Pierre Sermanet, Scott Reed, Dragomir
  Anguelov, Dumitru Erhan, Vincent Vanhoucke, and Andrew Rabinovich.
\newblock Going deeper with convolutions.
\newblock In \emph{Proceedings of the IEEE conference on computer vision and
  pattern recognition}, pages 1--9, 2015.

\bibitem[Tan and Le(2019)]{tan2019efficientnet}
Mingxing Tan and Quoc~V Le.
\newblock Efficientnet: Rethinking model scaling for convolutional neural
  networks.
\newblock \emph{arXiv preprint arXiv:1905.11946}, 2019.

\bibitem[Tan and Le(2021)]{tan2021efficientnetv2}
Mingxing Tan and Quoc~V Le.
\newblock Efficientnetv2: Smaller models and faster training.
\newblock 2021.

\bibitem[Tan et~al.(2019)Tan, Chen, Pang, Vasudevan, Sandler, Howard, and
  Le]{tan2019mnasnet}
Mingxing Tan, Bo~Chen, Ruoming Pang, Vijay Vasudevan, Mark Sandler, Andrew
  Howard, and Quoc~V Le.
\newblock Mnasnet: Platform-aware neural architecture search for mobile.
\newblock In \emph{Proceedings of the IEEE Conference on Computer Vision and
  Pattern Recognition}, pages 2820--2828, 2019.

\bibitem[Tang et~al.(2020)Tang, Wang, Xu, Chen, Shi, Xu, Xu, Tian, and
  Xu]{tang2020semi}
Yehui Tang, Yunhe Wang, Yixing Xu, Hanting Chen, Boxin Shi, Chao Xu, Chunjing
  Xu, Qi~Tian, and Chang Xu.
\newblock A semi-supervised assessor of neural architectures.
\newblock In \emph{Proceedings of the IEEE/CVF Conference on Computer Vision
  and Pattern Recognition}, pages 1810--1819, 2020.

\bibitem[Thrun and Pratt(1998)]{thrun98}
Sebastian Thrun and Lorien Pratt, editors.
\newblock \emph{Learning to Learn}.
\newblock Kluwer Academic Publishers, Norwell, MA, USA, 1998.
\newblock ISBN 0-7923-8047-9.

\bibitem[Vinyals et~al.(2016)Vinyals, Blundell, Lillicrap, Wierstra,
  et~al.]{vinyals2016matching}
Oriol Vinyals, Charles Blundell, Timothy Lillicrap, Daan Wierstra, et~al.
\newblock Matching networks for one shot learning.
\newblock In \emph{Advances in neural information processing systems (NIPS)},
  2016.

\bibitem[Wagstaff et~al.(2019)Wagstaff, Fuchs, Engelcke, Posner, and
  Osborne]{wagstaff2019setproof}
Edward Wagstaff, Fabian Fuchs, Martin Engelcke, Ingmar Posner, and Michael~A.
  Osborne.
\newblock On the limitations of representing functions on sets.
\newblock In \emph{Proceedings of the 36th International Conference on Machine
  Learning, {ICML} 2019, 9-15 June 2019, Long Beach, California, {USA}},
  volume~97 of \emph{Proceedings of Machine Learning Research}, pages
  6487--6494. {PMLR}, 2019.

\bibitem[Wang et~al.(2020)Wang, Wang, Yao, Shan, and Chen]{wang2020crossmodal3}
Sijin Wang, Ruiping Wang, Ziwei Yao, Shiguang Shan, and Xilin Chen.
\newblock Cross-modal scene graph matching for relationship-aware image-text
  retrieval.
\newblock In \emph{{IEEE} Winter Conference on Applications of Computer Vision,
  {WACV} 2020, Snowmass Village, CO, USA, March 1-5, 2020}, pages 1497--1506.
  {IEEE}, 2020.

\bibitem[Woodworth et~al.(2017)Woodworth, Gunasekar, Ohannessian, and
  Srebro]{woodworth2017learning}
Blake Woodworth, Suriya Gunasekar, Mesrob~I. Ohannessian, and Nathan Srebro.
\newblock Learning non-discriminatory predictors, 2017.

\bibitem[Wu et~al.(2019{\natexlab{a}})Wu, Dai, Zhang, Wang, Sun, Wu, Tian,
  Vajda, Jia, and Keutzer]{wu2019fbnet}
Bichen Wu, Xiaoliang Dai, Peizhao Zhang, Yanghan Wang, Fei Sun, Yiming Wu,
  Yuandong Tian, Peter Vajda, Yangqing Jia, and Kurt Keutzer.
\newblock Fbnet: Hardware-aware efficient convnet design via differentiable
  neural architecture search.
\newblock In \emph{Proceedings of the IEEE Conference on Computer Vision and
  Pattern Recognition}, pages 10734--10742, 2019{\natexlab{a}}.

\bibitem[Wu et~al.(2019{\natexlab{b}})Wu, Dai, Zhang, Wang, Sun, Wu, Tian,
  Vajda, Jia, and Keutzer]{wu_fbnet_2019}
Bichen Wu, Xiaoliang Dai, Peizhao Zhang, Yanghan Wang, Fei Sun, Yiming Wu,
  Yuandong Tian, Peter Vajda, Yangqing Jia, and Kurt Keutzer.
\newblock {FBNet}: {Hardware}-{Aware} {Efficient} {ConvNet} {Design} via
  {Differentiable} {Neural} {Architecture} {Search}.
\newblock \emph{arXiv:1812.03443 [cs]}, May 2019{\natexlab{b}}.
\newblock URL \url{http://arxiv.org/abs/1812.03443}.
\newblock arXiv: 1812.03443.

\bibitem[Xie et~al.(2017)Xie, Girshick, Doll{\'a}r, Tu, and
  He]{xie2017aggregated}
Saining Xie, Ross Girshick, Piotr Doll{\'a}r, Zhuowen Tu, and Kaiming He.
\newblock Aggregated residual transformations for deep neural networks.
\newblock In \emph{Proceedings of the IEEE conference on computer vision and
  pattern recognition}, pages 1492--1500, 2017.

\bibitem[Xiong et~al.(2021)Xiong, Xiong, Li, Tang, Liu, Bennett, Ahmed, and
  Overwikj]{xiong2021ir2}
Lee Xiong, Chenyan Xiong, Ye~Li, Kwok-Fung Tang, Jialin Liu, Paul~N. Bennett,
  Junaid Ahmed, and Arnold Overwikj.
\newblock Approximate nearest neighbor negative contrastive learning for dense
  text retrieval.
\newblock In \emph{International Conference on Learning Representations}, 2021.

\bibitem[Xu et~al.(2019)Xu, Hu, Leskovec, and Jegelka]{xu2018how}
Keyulu Xu, Weihua Hu, Jure Leskovec, and Stefanie Jegelka.
\newblock How powerful are graph neural networks?
\newblock In \emph{International Conference on Learning Representations}, 2019.

\bibitem[Xu et~al.(2020)Xu, Xie, Zhang, Chen, Qi, Tian, and Xiong]{xu2020pc}
Yuhui Xu, Lingxi Xie, Xiaopeng Zhang, Xin Chen, Guo-Jun Qi, Qi~Tian, and
  Hongkai Xiong.
\newblock Pc-darts: Partial channel connections for memory-efficient
  architecture search.
\newblock In \emph{International Conference on Learning Representations
  (ICLR)}, 2020.

\bibitem[Yan et~al.(2020{\natexlab{a}})Yan, Gong, Wei, and Gao]{yan2020deep}
Chenggang Yan, Biao Gong, Yuxuan Wei, and Yue Gao.
\newblock Deep multi-view enhancement hashing for image retrieval.
\newblock \emph{IEEE Transactions on Pattern Analysis and Machine
  Intelligence}, 2020{\natexlab{a}}.

\bibitem[Yan et~al.(2020{\natexlab{b}})Yan, Zheng, Ao, Zeng, and
  Zhang]{yan2020does}
Shen Yan, Yu~Zheng, Wei Ao, Xiao Zeng, and Mi~Zhang.
\newblock Does unsupervised architecture representation learning help neural
  architecture search?
\newblock \emph{Advances in Neural Information Processing Systems}, 33,
  2020{\natexlab{b}}.

\bibitem[Yurochkin et~al.(2020)Yurochkin, Bower, and
  Sun]{Yurochkin2020Training}
Mikhail Yurochkin, Amanda Bower, and Yuekai Sun.
\newblock Training individually fair ml models with sensitive subspace
  robustness.
\newblock In \emph{International Conference on Learning Representations}, 2020.
\newblock URL \url{https://openreview.net/forum?id=B1gdkxHFDH}.

\bibitem[Zaheer et~al.(2017)Zaheer, Kottur, Ravanbakhsh, Poczos, Salakhutdinov,
  and Smola]{deepset}
Manzil Zaheer, Satwik Kottur, Siamak Ravanbakhsh, Barnabas Poczos, Ruslan~R
  Salakhutdinov, and Alexander~J Smola.
\newblock Deep sets.
\newblock In \emph{Advances in Neural Information Processing Systems (NIPS)},
  2017.

\bibitem[Zemel et~al.(2013)Zemel, Wu, Swersky, Pitassi, and
  Dwork]{pmlr-v28-zemel13}
Rich Zemel, Yu~Wu, Kevin Swersky, Toni Pitassi, and Cynthia Dwork.
\newblock Learning fair representations.
\newblock In Sanjoy Dasgupta and David McAllester, editors, \emph{Proceedings
  of the 30th International Conference on Machine Learning}, volume~28 of
  \emph{Proceedings of Machine Learning Research}, pages 325--333, Atlanta,
  Georgia, USA, 17--19 Jun 2013. PMLR.
\newblock URL \url{https://proceedings.mlr.press/v28/zemel13.html}.

\bibitem[Zhang et~al.(2018)Zhang, Lemoine, and Mitchell]{zhang2018mitigating}
Brian~Hu Zhang, Blake Lemoine, and Margaret Mitchell.
\newblock Mitigating unwanted biases with adversarial learning, 2018.

\bibitem[Zhang et~al.(2019)Zhang, Jiang, Cui, Garnett, and Chen]{zhang2019d}
Muhan Zhang, Shali Jiang, Zhicheng Cui, Roman Garnett, and Yixin Chen.
\newblock D-vae: A variational autoencoder for directed acyclic graphs.
\newblock In \emph{Advances in Neural Information Processing Systems
  (NeurIPS)}, 2019.

\bibitem[Zhang et~al.(2020)Zhang, Zhao, Zhao, Yin, Yang, and
  Beutel]{zhang2020deep}
Weinan Zhang, Xiangyu Zhao, Li~Zhao, Dawei Yin, Grace~Hui Yang, and Alex
  Beutel.
\newblock Deep reinforcement learning for information retrieval: Fundamentals
  and advances.
\newblock In \emph{Proceedings of the 43rd International ACM SIGIR Conference
  on Research and Development in Information Retrieval}, pages 2468--2471,
  2020.

\bibitem[Zhen et~al.(2019)Zhen, Hu, Wang, and Peng]{zhen2019crossmodal1}
Liangli Zhen, Peng Hu, Xu~Wang, and Dezhong Peng.
\newblock Deep supervised cross-modal retrieval.
\newblock In \emph{{IEEE} Conference on Computer Vision and Pattern
  Recognition, {CVPR} 2019, Long Beach, CA, USA, June 16-20, 2019}, pages
  10394--10403. Computer Vision Foundation / {IEEE}, 2019.

\bibitem[Zhou et~al.(2020)Zhou, Zhou, Zhang, Loy, Yi, Zhang, and
  Ouyang]{zhou2020econas}
Dongzhan Zhou, Xinchi Zhou, Wenwei Zhang, Chen~Change Loy, Shuai Yi, Xuesen
  Zhang, and Wanli Ouyang.
\newblock Econas: Finding proxies for economical neural architecture search.
\newblock In \emph{Proceedings of the IEEE/CVF Conference on Computer Vision
  and Pattern Recognition}, pages 11396--11404, 2020.

\bibitem[Zoph and Le(2017)]{zoph2017rl}
Barret Zoph and Quoc~V. Le.
\newblock Neural architecture search with reinforcement learning.
\newblock In \emph{International Conference on Learning Representations
  (ICLR)}, 2017.

\bibitem[Zoph et~al.(2018)Zoph, Vasudevan, Shlens, and Le]{zoph2018learning}
Barret Zoph, Vijay Vasudevan, Jonathon Shlens, and Quoc~V Le.
\newblock Learning transferable architectures for scalable image recognition.
\newblock In \emph{Proceedings of the IEEE conference on computer vision and
  pattern recognition}, pages 8697--8710, 2018.

\end{thebibliography}

\clearpage

\appendix

\section*{Appendix}

\title{Supplementary File for Submission $\#$1191: \\Task-Adaptive Neural Network Retrieval \\ with Meta-Contrastive Learning}

\paragraph{Organization} In Appendix, we provide detailed descriptions of the materials that are not fully covered in the main paper, and provide additional experimental results, which are organized as follows:
\vspace{-0.05in}
\begin{itemize}
    \item \textbf{Section~\ref{suppl:implementation}} - We describe the \textit{implementation details} of our model-zoo construction, query and model encoders, and meta-surrogate performance predictor.
    \item \textbf{Section~\ref{suppl:train-details}} - We provide the details of the \textit{model training}, such as the learning rate and hyper-parameters, of meta-train/test and constructing model-zoo.
    \item \textbf{Section~\ref{suppl:proof}} - We provide the \textit{proof of injectiveness} with the proposed query and model encoding functions over the cross-modal latent space.
    \item \textbf{Section~\ref{suppl:experiment}} - We elaborate on the detailed \textit{experimental setups}, such as architecture space, baselines, and datasets, corresponding to the experiments introduced in the main document.
    \item \textbf{Section~\ref{suppl:more_exps}} - We provide additional \textit{analysis} of the experiments introduced in the main document and present the experiment with different model-zoo settings.
    \item \textbf{Section~\ref{sec:discuss}} - We discuss \textit{the societal impact and the limitation} of our work.

\end{itemize}


\section{Implementation Details}
\label{suppl:implementation}

\subsection{Efficient Model Zoo Construction}
\label{suppl:modelzoo_construction}

\begin{algorithm}[H]
    \begin{spacing}{0.8}
    \SetAlgoLined
    \SetKwInOut{Input}{Input}\SetKwInOut{Output}{Output}
    \SetKwRepeat{Repeat}{repeat}{until}
    \SetInd{0.5em}{2em}
        \Input{
            $\mathcal D, \mathcal M$: collection of datasets and models, respectively,\\
            $\mathcal Z_{(0)} \subset \mathcal D \times \mathcal M \times \mathbb R_{[0, 1]}$: set of $N_{init}$ initial tuples of (dataset, model, test accuracy)
        }
        $t \leftarrow 0$\;
        \While{termination condition is not met}{
            \If{$t$ is divisible by $N_{train}$}{
                Train accuracy predictor parameters $\bm\psi_{zoo}$ on data $\mathcal Z_{(t)}$\;
            }
            $C_{(t)} \leftarrow\ $ Choose a subset of candidate pairs from $\mathcal D \times \mathcal M$ not present in $\mathcal Z_{(t)}$\;
            $(D, M) \leftarrow \max_{(D, M) \in C_{(t)}} f_{zoo}(C; \mathcal Z_{(t)})$\;
            $\alpha^* \leftarrow$ Evaluate the actual accuracy of $(D, M)$ by training $M$ on $D$\; 
            $\mathcal Z_{(t+1)} \leftarrow \mathcal Z_{(t)} \cup (D, M, \alpha^{*})$\;
            $t \leftarrow t + 1$\;
        }
        \Output{Efficiently constructed model zoo $\mathcal Z_{(t)}$.}
        \caption{Model Zoo Construction}
        \label{alg:modelzoo}
    \end{spacing}
\end{algorithm}

The algorithm that is used to efficiently construct the model-zoo is described in Algorithm~\ref{alg:modelzoo}. The score function $f_{zoo}(D,M; \mathcal{Z})$, which measures how much adding a pair $(D, M)$ will improve upon the model zoo $\mathcal Z$, is defined as
\begin{align}
    f_{zoo}(D,M; \mathcal{Z}) := \mathbb{E}_{\hat{s}_{acc} \sim S((D,M); {\bm{\psi}_{zoo}})}[g_D(\mathcal{Z} \cup (D, M, \hat{s}_{acc})) - g_D(\mathcal{Z})] \label{eq:fzoo}
\end{align}
where $S$ indicates the accuracy predictor, and $g_D$ is the normalized volume under the pareto-dominated pairs for dataset $D$:
\begin{align}
    &g_D\left(\left\{ (D^{(i)}, M^{(i)}, s_{acc}^{(i)}) \right\}_{i=1}^{n} \right) \\
    :=&\; \mathrm{Hypervolume}\left(\left\{\left(s_{acc}^{(i)}, \tilde s_{latency}(M^{(i)}), \tilde s_{parameters}(M^{(i)})\right) \,\middle\vert\, D^{(i)} = D\right\}\right)
\end{align}
where $\tilde s_{latency}(M)$ and $\tilde s_{parameters}(M)$ indicates the normalized latency and the normalized number of parameters of the model $M$, respectively. The latency and parameters are normalized so that the maximum value across all models becomes 1.0 and the minimum value becomes 0.0. The hypervolume can be computed efficiently with the \texttt{PyGMO} library~\cite{Biscani2020}.

The accuracy predictor used in the model-zoo construction is very similar to the structure of the performance predictor described in Section~\ref{sec:perf_predictor}, but we used a functional embedding obtained from the model pretrained on Imagenet-1K, instead of a functional embedding obtained from a model already trained on the target dataset, since training the model on the target dataset just to obtain the functional embedding would defeat the purpose of this algorithm. Also, to incorporate uncertainty about the accuracy predictions, we use 10 samples from the accuracy predictor with MC dropout to evaluate the expectation in (\ref{eq:fzoo}). The dropout probability is set to 0.5.

\subsection{Query Encoder} 
Our query encoder takes sampled instances (e.g. 10 unseen random images per class) as an input from the query dataset. We use image embeddings from ResNet18~\cite{he2016resnet} pretrained on ImageNet 1K~\cite{deng2009imagenet}, whose dimensions are 512 (except for the last classification layer), rather than using raw images, simply to reduce computation costs. We then use a linear layer with 512 dimensions, followed by $Mean$ pooling and $L_2$ normalization, which outputs encoded vectors with $128$ dimensions. As we use Deep set~\cite{deepset}, we tried performing $Sum$ pooling, instead of $Mean$ pooling, however, we observe that taking the average on instances shows better R@$k$ scores for the correct pair retrieval, and thus we use $Mean$ pooling when encoding query samples.

\subsection{Model Encoder} 
Our model encoder takes both OFA flat topology~\cite{cai2020once} and functional embedding as an input. For the flat topology, it contains information such as kernel size, width expansion ratio, and depth, in a 45-dimensional vector. In addition, the functional embedding, which bypasses the need for direct parameter encoding, represents models' learned knowledge in a 1536-dimensional vector. We first concatenate both vectors and normalize the vector. Then, we learn a projection layer, which is a 1581-length fully-connected layer, followed by $L_2$ norm operation, which outputs encoded vectors with $128$ dimensions.

\subsection{Meta-Surrogate Performance Predictor}
\label{sec:perf_predictor}

Our performance predictor takes both query and model embeddings simultaneously. Both embedding vectors are 128-dimensional vectors. We first concatenate the embeddings into 256-dimensional vectors and then forward them through a 256-length fully connected layer. We then produce a single continuous value for a predicted accuracy. We perform a sigmoid operation on the values to range the values into a scale from 0.0 to 1.0.

%

\section{Training Details}
\label{suppl:train-details}

There are two steps of training required for our Task-Adaptive Network Search (TANS): 1) training the cross-modal latent space and 2) fine-tuning the retrieved model on an unseen meta-test dataset.  

\subsection{Learning the Cross-modal Latent Space}
For the model-zoo encoding, we set the batch size to 140 as we have 140 different datasets. Since, for each dataset, we randomly choose one model among 100 models from each dataset. Then we minimize the contrastive loss on the 140 samples. Although we train our encoders over a large number of dataset-network pairs (14,000 models), the entire training time takes less than two hours based on NVIDIA's RTX 2080 TI GPU. We initialize our model weights with the same value across all encoders and experiments, rather than differently initializing the encoders for every experimental trial. We use the Adam optimizer (We use the learning rate of $1$e-$2$).

\subsection{Fine-tuning on Meta-test Datasets}
For the fine-tuning phase, we set all settings, such as hyper-parameters, learning rate, optimizer, etc., exactly the same across all baseline models and our method, and the differences are clearly mentioned in this section otherwise. We use the SGD optimizer with an initial learning rate ($1$e-$2$), weight decay ($4$e-$5$), and momentum ($0.9$). Also, we use the \textit{Cosine Annealing} learning rate scheduler. We train the models with $224$ sized images (after resizing) and we set batch-size to $32$, except PC-DARTS which has memory issues with $224$ sized images (for PC-DARTS, we set to $12$ as the batch-size), and DrNAS which we train with $32$ sized images due to \textit{heavy} training time costs. We train all models for $50$ epochs and we show that our model converges faster than all baseline models.

\subsection{Constructing the Model-Zoo}

For the model-zoo consisting of 14,000 random pairs used in the main experiment, we fine-tune the ImageNet1K pretrained OFA models on the dataset for 625 epochs, following the progressive shrinking method described in \cite{cai2020once}. We then choose 100 random OFA architectures for each dataset and evaluate their test accuracies on the test split.

For the efficiently constructed model zoo experiment, we use the algorithm described in Section 3.3 and further elaborated in Section~\ref{suppl:modelzoo_construction}, using the 14,000-pair model zoo as the search space. For the initial samples, we use $N_{init} = 750$, where 5-6 samples are taken from each dataset. The accuracy predictor is retrained from scratch every 64 iterations until the validation accuracy no longer improves for 5 epochs.


\section{Proof for Uniqueness of the Query and Model Encoding Functions}
\label{suppl:proof}

In this section, we show that the proposed query and model encoders can represent the injective function on the input query $D \in \mathcal{Q}$ and model $M \in \mathcal{M}$, respectively.

\textbf{Proposition 1 (Injectiveness on Query Encoding).} \emph{Assume $\mathcal{Q}$ and $D$ are finite sets. A query encoder $E_Q: \mathcal{Q} \rightarrow \mathbb{R}^d$ can injectively map two different queries $D_1, D_2$ into distinct embeddings $\bm{q_1}, \bm{q_2}$, where $D \in \mathcal{Q}$ and $\bm{q} \in \mathbb{R}^d$.}

\begin{proof}
A query encoder $E_Q$ maps a query dataset $D \in \mathcal{Q}$ to a vector $\bm{q} \in \mathbb{R}^d$ as follows: $E_Q: D \mapsto \bm{q}$, where $\mathcal{Q}$ is a set of queries, which contains a set of data instances $X$ for constructing a dataset $D = \left\{ X_1, X_2, ..., X_n \right\}$. Then, our goal here is to make a query encoder that uniquely maps two different queries $D_1, D_2$ into two distinct embeddings $\bm{q}_1, \bm{q}_2$.

Each dataset $D$ consists of $n$ data instances: $D = \left\{ X_1, X_2, ..., X_n \right\}$, where $n$ is smaller than the number of elements in $\mathbb{N}$. To encode each query dataset $D$ into a vector space, as described in query encoder paragraph of section 3.2, we first transform each instance $X_i$ into the representation space with a continuous function $\rho$, and then aggregate all set elements, which is adopted from~\citet{deepset}. In other words, a query encoder can be defined as follows: $\bm{q} = \sum_{X_i \in D} \rho(X_i)$.

We assume that $\mathcal{Q}$ is a finite set, and each $D = \left\{ X_1, X_2, ..., X_n \right\} \in \mathcal{Q}$ is also a finite set with $\left| D \right| = n$ elements. Therefore, a set of data instances $X$ is countable, since the product of two nature numbers (i.e. $\left| \mathcal{Q} \right| \times n$) is a natural number. For this reason, there can be a unique mapping $Z$ from the element $X$ to the nature number in $\mathbb{N}$. If we let $\rho(X) = 4^{-Z(X)}$, then the form of a query encoder $\sum_{X_i \in D} \rho(X_i)$ constitutes an unique mapping for every set $D \in \mathcal{Q}$ (see \citet{deepset, wagstaff2019setproof} for details). In other words, the output of the query encoder is unique for each input dataset $D$ that consists of $n$ data instances.

\end{proof}

Thanks to the universal approximation theorem~\cite{Hornik89universal, hornik91universal}, we can construct a mapping function $\rho$ using multi-layer perceptrons (MLPs).

\textbf{Proposition 2 (Injectiveness on Model Encoding).} \emph{Assume $\mathcal{M}$ is a countable set. A model encoder $E_M: \mathcal{M} \rightarrow \mathbb{R}^d$ can injectively map two different architectures $M_1, M_2$ into distinct embeddings $\bm{m}_1, \bm{m}_2$, where $M \in \mathcal{M}$ and $\bm{m} \in \mathbb{R}^d$}.

\begin{proof}
As described in the model encoder paragraph of Section 3.2, we represent each neural network $M \in \mathcal{M}$ with both topological embedding $\bm{v}_t$ and functional embedding $\bm{v}_f$. Thus, if one of two embeddings can uniquely represent each neural network, then the injectiveness on model encoding $E_M: \mathcal{M} \rightarrow \mathbb{R}^d$ is satisfied.

We first show that the topological encoding function $E_{M_T}: M \mapsto \bm{v}_t$ can uniquely represent each architecture $M$ in the embedding space. As described in the Model Encoder part of section~\ref{suppl:implementation}, we use a 45-dimensional vector that contains topological information, such as the number of layers, channel expansion ratios, and kernel sizes (See \citet{cai2020once} for details), for the topological encoding. Also, each topological information uniquely defines each neural architecture. Therefore, the embedding $\bm{v}_t$ from the topological encoding function $E_{M_T}$ is unique on each neural network $M$.

While we can obtain the distinct embedding of each neural network with the topological encoding function alone, we also consider the injectiveness of the functional encoding in the following. To consider the functional embedding, we first model a neural architecture as its computational graph, which can be further denoted as a directed acyclic graph (DAG). Using this computational graph scheme, a functional model encoder $E_{M_F}$ maps an architecture (computational graph) $M \in \mathcal{M}$ into a vector $\bm{v}_f$ as follows: $E_{M_F}: M \mapsto \bm{v}_f$. Then, our goal here is to make the functional encoder $E_{M_F}$ that uniquely maps two different neural architectures $M_1, M_2$ into two different embeddings ${\bm{v}_f}_1, {\bm{v}_f}_2$, with the computational graph represented as the DAG structure. 

Assume that a computational graph for a neural network $M$ has $n$ nodes. Then, each node $v_i$ on the graph has its corresponding operation $o_i$, which transforms incoming features for the node $v_i$ to an output representation $C_i$. In other words, $C_i$ indicates the output of the composition of all operations along the path from $v_1$ to $v_i$. 

Particularly, in our model encoder case, we have an arbitrary input signal $\bm{x}$ that is the fixed Gaussian noise, where we insert this fixed input into the starting node $v_1$ (See Model Encoder paragraph of section 3.2 for details). Also, for the simplicity of the notation, we set $C_0 = \bm{x}$ that is the output of the virtual node $v_0$ and the incoming representation of the starting node of the computational graph. Then, the output representation for the node $v_i$ is formally defined as follows: $C_i(\bm{x}) = o_i(\left\{ C_j(\bm{x}): v_j \rightarrow v_i \right\})$, where $\left\{ C_j(\bm{x}): v_j \rightarrow v_i \right\}$ denotes a multiset for the output representation of $v_i$'s predecessors, and the operation $o_i$ transforms the incoming representations over the multiset into the output representation. Note that, to consider the multiplicity of the nodes on a graph, we use a multiset scheme, rather than a set~\cite{xu2018how}.

Following the proof of Theorem 2 in~\citet{zhang2019d}, we rewrite the $C_i(\bm{x}) = o_i(\left\{ C_j(\bm{x}): v_j \rightarrow v_i \right\})$ as follows: $C_i(\bm{x}) = \omega(o_i, \left\{ C_j(\bm{x}): v_j \rightarrow v_i \right\})$, where $\omega$ is an injective function over two inputs $o_i$ and $\left\{ C_j(\bm{x}): v_j \rightarrow v_i \right\}$. Then, $C_i$ can uniquely embed the output representation of node $v_i$, and this is an injective (Please refer to the proof of Theorem 2 in~\citet{zhang2019d} for details). Thus, the output of the computational graph for the network $M$ with the fixed Gaussian input noise $\bm{x}$ is uniquely represented with the functional encoder $E_{M_F}: M \mapsto \bm{v}_f$, where $\bm{v}_f = C_n$ with $n$ nodes on the graph.


Note that we use a network $M$ that is task-adaptively trained for a specific target dataset to not only obtain high performance on the target dataset but also reduce the fine-tuning cost on it. Thus, while we might further need to consider the parameters on the computational graph, we show the injectiveness on the functional encoding only with the computational graph structure and leave the consideration of parameters as a future work, since it is complicated to formally define the injectiveness with trainable parameters.

To sum up, we show the injectiveness of the model representation with both topological encoding and functional encoding schemes, although only one encoding function can injectively represent the entire neural network. While we further concatenate and transform two output representations with a function $g$, to obtain the final model representation: $\bm{m} = g([\bm{v}_t, \bm{v}_f])$, the representation $\bm{m}$ is also unique on each neural network $M$ with an injective function $g$.

\end{proof}

Similar to the universal approximation theorem~\cite{Hornik89universal, hornik91universal}, we might construct an injective mapping function $g$ and $\omega$ with learnable parameters on it.


\begin{figure*}[t!]
\vspace{0.1in}
\centering
\hspace{0in} Query \hspace{0.75in} Retrieval \hspace{1in} Query \hspace{0.75in} Retrieval
\\
\subfloat{
\includegraphics[width=.23\linewidth]{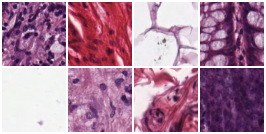}
\includegraphics[width=.23\linewidth]{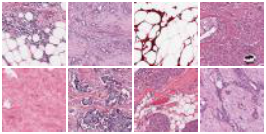}
}
\quad
\subfloat{
\includegraphics[width=.23\linewidth]{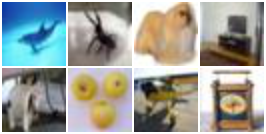}
\includegraphics[width=.23\linewidth]{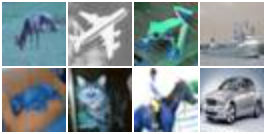}}
\\
\subfloat{
\includegraphics[width=.23\linewidth]{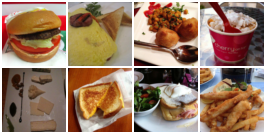}
\includegraphics[width=.23\linewidth]{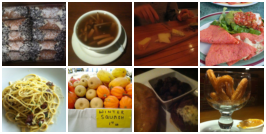}
}
\quad
\subfloat{
\includegraphics[width=.23\linewidth]{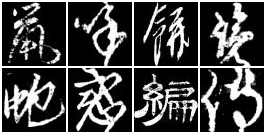}
\includegraphics[width=.23\linewidth]{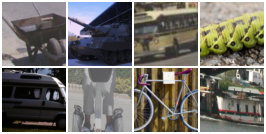}
}
\\
\subfloat{
\includegraphics[width=.23\linewidth]{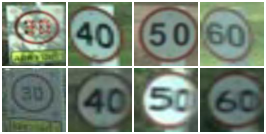}
\includegraphics[width=.23\linewidth]{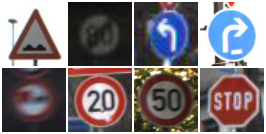}
}
\quad
\subfloat{
\includegraphics[width=.23\linewidth]{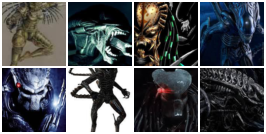}
\includegraphics[width=.23\linewidth]{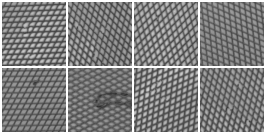}
}
\\
\subfloat{
\includegraphics[width=.23\linewidth]{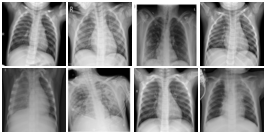}
\includegraphics[width=.23\linewidth]{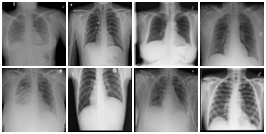}
}
\quad
\subfloat{
\includegraphics[width=.23\linewidth]{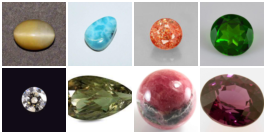}
\includegraphics[width=.23\linewidth]{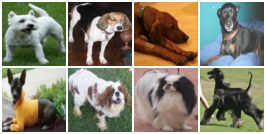}
}

\subfloat{
\includegraphics[width=.23\linewidth]{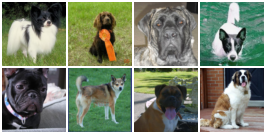}
\includegraphics[width=.23\linewidth]{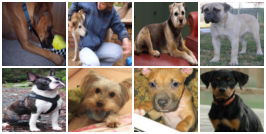}
}
\quad
\subfloat{
\includegraphics[width=.23\linewidth]{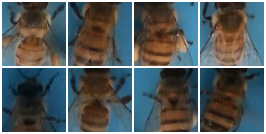}
\includegraphics[width=.23\linewidth]{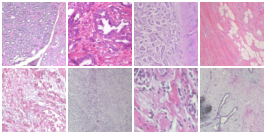}
}
\vspace{-0.05in}
\caption{\small 
\textbf{Retrieved Examples from 10 Meta-test Real-world Datasets.} We present all query-retrieval pairs on meta-test datasets. Each row includes two pairs of query dataset (left) and the retrieved dataset (right). Please see detailed explanations of the pairs in Section~\ref{suppl:dataset_details} and Table~\ref{tbl:dataset}. 
}
\label{fig:unseen_retrieval_10}
\vspace{-0.05in}
\end{figure*}

\section{Experimental Setup}
\label{suppl:experiment}

\subsection{Architecture Space}
\label{suppl:architecture_space}
Before constructing a model zoo that contains a large number of dataset-architecture pairs, we first need to define an architecture search space on it to handle all architectures in a consistent manner. To easily obtain the task-adaptive parameters for the given task with consideration of various factors, such as a number of layers, kernel sizes, and width expansion ratios, we use the supernet-based OFA architecture space~\cite{cai2020once}, which the same as the well-known MobileNetV3 space~\cite{howard2019searching}. Each neural architecture in the search space consists of a stack of 20 mobile-block convs (MBconvs), where the number of units is 5, and the number of layers on each unit ranges across $\{2,3,4\}$. Moreover, for each layer, we select the kernel size is from $\{3, 5, 7\}$, and the width-depth ratio from $\{3, 4, 6\}$. This strategy allows us to generate around $10^{19}$ neural architecture candidates in theory.

\subsection{Model Zoo}
\label{suppl:detail_model_zoo}

To construct a model zoo consisting of a large number of dataset-architecture pairs, we collect 89 real-world datasets for image classification from Kaggle\footnote[1]{\url{https://www.kaggle.com/}} and obtain 100 random architectures per dataset from the OFA space. Specifically, we first divide the collected datasets into two non-overlapping sets for meta-training and meta-testing. If the dataset has more than 20 classes, then we randomly split it into multiple datasets such that a dataset can consist of up to 20 classes. For meta-testing, we randomly selected only one of the splits for each original dataset for diversity. This process yields 140 datasets for meta-training and 10 datasets for meta-testing. To generate a validation set for each dataset, we randomly sample 20\% of data instances from each dataset and use the sampled instances for the validation, while using the remaining 80\% as the training instances. 

For statistics, the number of classes ranges from 2 to 20 with a median of 16, and the number of instances for each dataset ranges from 8 to 158K with a mean of 2,847. We then construct the model-zoo by fine-tuning 100 random OFA architectures on training instances of each dataset and obtaining their performances on its respective validation instances, which yields 14K (dataset, architecture, accuracy) tuples in total. We use this database throughout this paper. 

\subsection{Dataset Details}
\label{suppl:dataset_details} In Table~\ref{tbl:dataset}, we provide information of all datasets that we utilize for model-zoo construction and meta-test experiments, including the dataset name, a brief description of the dataset, the number of splits for train and test sets, and corresponding Kaggle URL. Please refer to the table if you look into a certain dataset more closely. Particularly, we further provide an explanation of all query and retrieval pairs in Figure~\ref{fig:unseen_retrieval_10}. Beginning from the left column on the first row, we present pairs of Colorectal Histology (query) \& Breast Histopathology (retrieval) and Real or Drawing (query) \& Tiny Images (retrieval). In the second row, we show pairs of Dessert (query) \& Food Kinds (retrieval) and Chinese Characters (query) \& Vehicles (retrieval). For the third row, there are pairs of Speed Limit Signs (query) \& Traffic Signs (retrieval) and Alien vs Predator (query) \& Grid Anomaly (retrieval). In the fourth row, we illustrate pairs of COVID-19 (query) \& Chest X-Rays (retrieval) and Gemstones (query) \& Stanford Dogs (retrieval). In the last row, we present pairs of Dog Breeds (query) \& Stanford Dogs (retrieval) and Honeybee Pollen (query) \& Breast Cancer Tissues (retrieval). Please see Table~\ref{tbl:dataset} for the detailed information for each dataset.

\subsection{Baseline NAS Methods} Here we describe the baselines we use in the experiments in the main document. We compare the performance of the models retrieved with our method against pretrained neural networks as well as the ones searched by several efficient NAS methods that are closely related to ours:

\textbf{1) MobileNetV3}~\citep{howard2019searching} MobileNetV3 is a representative resource-efficient neural architecture tuned considering mobile phone environments. In our experiments, MobileNetV3 is pretrained on ImageNet-1K, which is fine-tuned for 50 epochs on each meta-testing task.

\textbf{2) PC-DARTS}~\citep{xu2020pc}, a  differentiable NAS method based on a weight sharing scheme that reduces search time efficiently and especially improves memory usage, search time, performance compared to DARTS~\cite{liu2018darts} by designing partial channel sampling and edge normalization. We search for architectures for each meta-testing task by following the official code at \url{https://github.com/yuhuixu1993/PC-DARTS}.

\textbf{3) DrNAS}~\citep{chen2021drnas}, a differentiable NAS method that handles NAS as a distribution problem, modeled by Dirichlet distribution. We use the official code at \url{https://github.com/xiangning-chen/DrNAS}.

\textbf{4) OFA}~\citep{cai2020once}, a NAS method that provides a subnet sampled from a larger network (supernet) pretrained on ImageNet-1K, which alleviates the performance degeneration of prior supernet-based methods. 
We use the code at \url{https://github.com/mit-han-lab/once-for-all}. 

\textbf{5) MetaD2A}~\citep{lee2021rapid}, a meta-NAS model that rapidly generates data-dependent architecture for a given task that is meta-learned on subsets of ImageNet-1K. From the ImageNet-1K dataset and architectures of OFA search space, we randomly use 3296 and 14,000 meta-training tasks for the generator and predictor, respectively as a source database.

\textbf{6) FBNet}~\citep{wu_fbnet_2019}, a collection of convolutional models obtained via Differentiable Neural Architecture Search. We use FBNet-A pretrained on ImageNet-1K and fine-tune it on each meta-testing task for 50 epochs.

We use the same hyper-parameters for all baselines for a fair comparison. We fine-tune the architecture for 50 epochs on each meta-testing task. The SGD optimizer is used with a learning rate of 0.01, the momentum of 0.9, and 4e-5 weight decay. The image size is 224$\times$224 and the batch size is 32.

\section{Additional Experiments \& Analysis}
\label{suppl:more_exps}

\subsection{Experiment on Real-world Networks}
While the proposed TANS shows outstanding performances in a number of neural network search tasks with the manicured architecture search space described in Section~\ref{suppl:architecture_space}, it could be more beneficial if we search the best-fitted model on a given query dataset from the pretrained networks with real-world neural architectures (e.g. ResNet) trained on various datasets. For this even more realistic scenario for Neural Network Search (NNS), we construct our model-zoo including \textbf{ten real-world architectures}, such as ResNet18~\cite{he2016resnet}, ShuffleNet V2~\cite{ma2018shufflenet}, MobileNet v2~\cite{sandler2018mobilenetv2}, SqueezeNet~\cite{iandola2016squeezenet}, GoogLeNet~\cite{szegedy2015going}, ResNeXt~\cite{xie2017aggregated}, AlexNet~\cite{krizhevsky2012imagenet}, MNASNet~\cite{tan2019mnasnet}, EfficientNet-B0~\cite{tan2019efficientnet}, and LambdaResNet~\cite{bello2021lambdanetworks}. 

\paragraph{Experimental Setup} To construct the new \textbf{real-world model-zoo}, we first meta-train the real-world architectures and merge the new dataset-network pairs (1,400 pairs) with the random subset of the previous model-zoo (about 5,000 OFA-based models), yielding about 6,500 models in the new model-zoo. The way of training is the same as the experiment introduced in the main document (Section 4.1) except that we only use functional embeddings, while topology information is not used when learning the cross-modal latent space (we exclude the topology information since encoding the topologies of real networks across various search spaces into a single uniform format is too complicated.) Including the real-world architectures, we first verify the retrieval performance of our model on the meta-train datasets, and our TANS achieves 90 for the R@1, 100 for the R@5 scores. The way of evaluating on the meta-test dataset is also the same as the experiments that we conducted in the main document (10 real-world meta-test datasets), except that we conduct experiments only on five datasets out of the ten datasets used in the experiments of the main document, due to the heavy training costs required for meta-testing. The selected datasets are Speed Limit Signs, Honey-bee Pollen, Alien-vs-Predator, Chinese Characters, and COVID-19 datasets (for detailed information for each dataset, please see Table~\ref{suppl:dataset_details}.) 

\paragraph{Experimental Results}  In Table~\ref{tbl:real-world}, our methods, both with OFA and the real-world architectures, outperform all baseline models, including MobileNetV3 (about $1.0\%p$ to $1.5\%p$ higher), PC-DARTS (about $17.5\%p$ to $18.0\%p$ higher), DrNAS (about $14.5\%p$ to $15.0\%p$ higher), FBNet (about $1.5\%p$ to $2.0\%p$ higher), OFA (about $1.5\%p$ to $2.0\%p$ higher), and MetaD2A (about $0.5\%p$ to $1.0\%p$ higher). We observe that collecting more lightweight real-world neural network and dataset pairs (TANS w/ Real-world Model-Zoo) will allow our model to retrieve computationally efficient pretrained networks in a task-adaptive manner. Such data-driven nature is another advantage of our method since we can easily increase the performance of the model by collecting more pretrained networks that are readily available in many public databases. 


\begin{table*}[t!]
	\small
	\vspace{-0.1in}
	\caption{\small \textbf{Performance Comparison on 5 Unseen Real-world Datasets} All reported results are average performances over 3 different runs with $95\%$ confidence intervals.
	} 
	\vspace{-0.1in}
	\begin{center}
     \resizebox{\textwidth}{!}{
        \renewcommand{\arraystretch}{0.7}
	    \begin{tabular}{clccrcc}
	   	\toprule
	   	\multirow{2}{*}{Target Dataset}                              &
	   	\multicolumn{1}{c}{\multirow{2}{*}{Method}}                  & 
	   	Params                                  &
	   	Search Time &
	   	Training Time &

	   	\multirow{2}{*}{Speed Up}                                    &
	   	Accuracy                               \\
	   	& &(M)&(GPU sec)&(GPU sec)&&(\%)\\
	   	
	   	\midrule
	   	\midrule
	   	
        \multirow{7}{*}{\shortstack{\\ \\ \\ Averaged \\ Performance }} 
        &
	   	MobileNetV3~\citep{howard2019searching}          &
        4.00 &
        - &
        178.45\tiny$\pm$06.18 &
        1.00$\times$ &
        96.86\tiny$\pm$0.47    \\
        
        \cdashline{2-7}\noalign{\vskip 0.75ex}
	   	
        &
	   	PC-DARTS~\citep{xu2020pc} - 500 Epochs                      &
        \textbf{3.45} &
        943.17\tiny$\pm$15.44 &
        4255.74\tiny$\pm$1366.92 &
        0.03$\times$ &
        80.48\tiny$\pm$14.33              \\
	   	
	   		   	                                                &
	   DrNAS~\citep{chen2021drnas} - 500 Epochs                          &
        4.12 &
        873.44\tiny$\pm$25.78 &
        2445.22\tiny$\pm$76.54 &
        0.05$\times$ &
        83.58\tiny$\pm$2.79             \\
        
        \cdashline{2-7}\noalign{\vskip 0.75ex}
        
	   	                                       &
	    FBNet-A~\citep{wu_fbnet_2019}                       &
        4.30 &
        - &
        218.40\tiny$\pm$42.79&
        0.82$\times$&
        96.15\tiny$\pm$2.51                \\
        
	    &
	    OFA~\citep{cai2020once}                       &
        6.74 &
        121.90\tiny$\pm$0.00 &
        162.33\tiny$\pm$03.35 &
        0.63$\times$ &
        96.04\tiny$\pm$1.00                \\
                 
	   	&
	   	MetaD2A~\citep{lee2021rapid}                     &
        6.15 &
        2.56\tiny$\pm$0.15 &
        228.87\tiny$\pm$26.64 &
        0.77$\times$ &
        97.34\tiny$\pm$1.10                 \\
        
        \cdashline{2-7}\noalign{\vskip 0.75ex}
        
	   	&
	   	\textbf{TANS (Ours) } w/ OFA-Based Model-Zoo &
	   	5.50 &
	   	0.002\tiny$\pm$0.00 &
	   	121.18\tiny$\pm$10.71  &
	   	1.47$\times$ & 
	   	97.79\tiny$\pm$0.28 \\
	   	
	   	&
	   	\textbf{TANS (Ours) } w/ Real-world Model-Zoo & 
	   	5.43 &
	   	\textbf{0.001\tiny$\pm$0.00} &
	   	\textbf{115.06\tiny$\pm$16.82} &
	   	\textbf{1.55$\times$} &
	   	\textbf{98.59\tiny$\pm$0.38} \\

	   	\bottomrule
        \end{tabular}
	    }
	\end{center}
	\vspace{-0.1in}
	\label{tbl:real-world}
	\small
	\vspace{-0.1in}
\end{table*}

\subsection{Additional Performance Comparison with NAS Methods}
\begin{wraptable}[9]{r}{6cm}
\vspace{-0.2in}
    \small
    \centering
    \caption{Comparison with NAS methods}
    \vspace{-0.05in}
    \resizebox{0.4\textwidth}{!}{
    \begin{tabular}{l|ccc}
	   \toprule 
	   \centering
	   Method & \multicolumn{2}{c}{Meta-test Datasets} \\
	   & Colorectal & Food \\
      \midrule
      \midrule
      MetaD2A  &
      $96.57\%$ &  $89.72\%$ \\
      
      DrNAS w/ ImageNet    &
      $84.27\%$ &  $75.90\%$ \\
      
      PC-DARTS w/ ImageNet  &
      $96.77\%$ &  $86.75\%$ \\
      
      \textbf{TANS 1/10 (Ours)} &
      96.83$\%$ &  \textbf{94.31$\%$} \\
      
      \textbf{TANS (Ours)} &
      \textbf{97.67$\%$} &  93.71$\%$ \\
	  
	  \bottomrule
	   \end{tabular}	
    }
    \label{tbl:supp_more_exp}
\vspace{-0.1in}
\end{wraptable}
In the experiment introduced in the main document (Table~\ref{tbl:unseen_task}), we train DrNAS and PC-DARTS, which only generate architectures without pretrained weights, for 10 times more iterations (500 epochs) for a fair comparison (while the other methods, which share ImageNet pretrained knowledge, are trained for 50 epochs). In this experiment, rather than training for 500 epochs, we pretrain networks obtained by DrNAS and PC-DARTS on \textbf{``ImageNet''} and then fine-tune on two meta-test datasets (Colorectal Histology \& Food Classification Datasets). As shown in Table~\ref{tbl:supp_more_exp}, although pretraining on ImageNet improves their results, our methods, including TANS with 1/10 sized model-zoo (1400), still outperforms all baselines, which shows that retrieving and utilizing \textbf{pretrained weights of relevant tasks} is more effective than using ImageNet pre-trained weights.

\subsection{Synergistic Effect of TANS and State-of-the-Art NAS Methods}
Not only the real-world architectures but also any existing NAS methods can be successfully integrated with our retrieval framework by simply adding searched networks into our model-zoo. We demonstrate such synergistic effect of TANS and \textbf{NAS methods} in Figure~\ref{fig:convergence_and_params_acc} (e) of the main document. Constructing the model-zoo with neural architectures generated by MetaD2A, which is a state-of-the-art NAS method, improves our performance compared to the previous model-zoo that are simply sampled from the OFA search space. Considering that NAS approaches have been actively studied~\cite{lee2021rapid, tan2021efficientnetv2, brock2021high, tan2019efficientnet, srinivas2021bottleneck, dosovitskiy2021an} and pretrained models are often shared via open-source, we believe that the TANS framework has powerful potential to continuously improve its performance by absorbing such new models into the model-zoo.

\section{Discussion}
\label{sec:discuss}

\subsection{Societal Impacts}
\label{suppl:societal}
Our framework, TANS, has the following beneficial societal impacts: (1) enhanced accessibility, (2) preservation of data privacy, and (3) the reduction of reproducing efforts. 

\paragraph{Enhanced accessibility} Since our Task-Adaptive Neural Network Search (TANS) framework allows \textit{anyone} to \textit{instantly retrieve} a full neural network that works well on the given task, by providing only a \textit{small} set of data samples, it can greatly enhance the accessibility of AI to users with little knowledge and backgrounds. Moreover, it does not require large computational resources, unlike existing NAS or AutoML frameworks, which further helps with its accessibility. Finally, to allow everyone to benefit from our task-adaptive neural network search framework, we will publicly release our model-zoo, which currently contains \textit{more than 15K models}, and open-source it. Then, anyone will be able to freely retrieve/update any models from our model-zoo. 

\paragraph{Preservation of data-privacy}
Our framework requires only a small set of sampled data instances to retrieve the task-adaptive neural network, unlike existing NAS/AutoML methods that require a large number of data instances to search optimal architectures for the target datasets. Thus, the data privacy is largely improved, and we can further allow the set encoding to take place on the client-side, rather than at the server. This will result in enhanced data privacy, as none of the raw data samples need to be submitted to the system. 

\paragraph{Reduction of reproducing efforts} Many ML researchers and engineers are wasting their time and labors, as well as the computational and monetary resources in reproducing existing models and fine-tuning them. TANS, since it instantly retrieves a task-relevant model from a model zoo that contains a large number of state-of-the-art networks pretrained on diverse real-world datasets, the users need not redesign networks or retrain them at excessive costs. Since we plan to populate the model zoo with more pretrained networks, the coverage of the dataset and architectures will become even broader as time goes on. Since training deep learning models often requires extremely large computing cost, which is costly in terms of energy consumption, and results in high carbon emissions, our method is also environment-friendly.

\subsection{Limitations}
\label{subsec:limit}
As a prerequisite condition, our method must have a model-zoo which contains pretrained models that can cover diverse tasks and perform well on each given task. There exists a chance that TANS could be affected by biased initialization if the meta-training pool contains biased pretrained models. To prevent this issue, we can use existing techniques that ensure fairness when constructing a model-zoo, which identify and discard inappropriate datasets or models. There have been various studies for alleviating unjustified bias in machine learning systems. Fairness can be classified into individual fairness, treating similar users similarly \citep{10.1145/2090236.2090255,Yurochkin2020Training}, and group fairness, measuring the statistical parity between subgroups, such as race or gender \citep{pmlr-v28-zemel13, louizos2017variational, Hardt2016EqualityOO}. Optimizing fair metrics during training is achieved by regularizing the covariance between sensitive attributes and model predictions~\citep{woodworth2017learning} and minimizing an adversarial ability to estimate sensitive attributes from model predictions~\citep{zhang2018mitigating}. At evaluation times,~\citep{agarwal2018reductions, cotter2018training} improves the generalizability for a fair classifier via two-player games. All these methods can be adopted when building our model-zoo.



\onecolumn

\begin{center}

\begingroup

\fontsize{8pt}{11.8pt}\selectfont

\setlength\LTleft{0pt}            

\begin{longtable}{cp{0.9in}p{1.5in}crrl}
\caption{\small {\textbf{Dataset Details}} Detailed information, such as dataset name, description, and download link, about all datasets that we utilize are described (Due to the space limit, we provide hyperlinks to the webpage for the datasets, rather than printing the full website links.) } 
\label{tbl:dataset} \\

\hline 
\multicolumn{1}{c}{\textbf{No.}} & \multicolumn{1}{c}{\textbf{Dataset Name}} & \multicolumn{1}{c}{\textbf{Brief Description}} & \multicolumn{1}{c}{\textbf{Instances}} & \multicolumn{1}{c}{\textbf{Cls.}} & \multicolumn{1}{c}{\textbf{Splits}} & \multicolumn{1}{c}{\textbf{URL}} 
\\ \hline 
\endfirsthead

\multicolumn{7}{c}%
{{\bfseries \tablename\ \thetable{} -- continued from previous page}} \\
\hline \multicolumn{1}{c}{\textbf{No.}} & \multicolumn{1}{c}{\textbf{Dataset Name}} & \multicolumn{1}{c}{\textbf{Brief Description}} & \multicolumn{1}{c}{\textbf{Instances}} & \multicolumn{1}{c}{\textbf{Cls.}} & \multicolumn{1}{c}{\textbf{Splits}} & \multicolumn{1}{c}{\textbf{URL}} \\ \hline 
\endhead 

\hline \multicolumn{7}{|r|}{\textit{Continued on next page}} \\ \hline
\endfoot

\hline \hline
\endlastfoot

MetaTrain-1&{Store Items}&
Classify store item images by color&
4984 / 624&
12&
1&
\href{https://kaggle.com/imoore/6000-store-items-images-classified-by-color}{Link}
\\
MetaTrain-2&{Big Cats}&
Classify big cats by species&
2875 / 360&
4&
1&
\href{https://kaggle.com/c/DL2020}{Link}
\\
MetaTrain-3&{Deepfake Detection}&
Deepfake detection&
12000 / 1500&
2&
1&
\href{https://kaggle.com/c/ads5035-01}{Link}
\\
MetaTrain-4&{Food Kinds}&
Classify kinds of food&
10580 / 1322&
11&
1&
\href{https://kaggle.com/c/ai2020f}{Link}
\\
MetaTrain-5&{Hair Color}&
Classify people by hair color&
2560 / 320&
4&
1&
\href{https://kaggle.com/c/aia-dl-mid}{Link}
\\
MetaTrain-6&{Apparels}&
Classify apparel images by kind and color&
9091 / 1137&
24&
2&
\href{https://kaggle.com/trolukovich/apparel-images-dataset}{Link}
\\
MetaTrain-7&{Manual Alphabet}&
Classify manual alphabet letters&
69600 / 8700&
29&
2&
\href{https://kaggle.com/grassknoted/asl-alphabet}{Link}
\\
MetaTrain-8&{Artworks}&
Classify artworks by artist&
6997 / 877&
51&
3&
\href{https://kaggle.com/ikarus777/best-artworks-of-all-time}{Link}
\\
MetaTrain-9&{Blood Cells}&
Identify blood cell types&
9954 / 1244&
4&
1&
\href{https://kaggle.com/paultimothymooney/blood-cells}{Link}
\\
MetaTrain-10&{Breast Cancer Tissues}&
Idenify breast cancer with microscope images&
6323 / 789&
8&
1&
\href{https://kaggle.com/ambarish/breakhis}{Link}
\\
MetaTrain-11&{Breast Histopathology}&
Identify breast cancer with sample images&
222018 / 27753&
2&
1&
\href{https://kaggle.com/paultimothymooney/breast-histopathology-images}{Link}
\\
MetaTrain-12&{Aerial Cactus}&
Identify cacti in aerial photos&
17199 / 2150&
2&
1&
\href{https://kaggle.com/irvingvasquez/cactus-aerial-photos}{Link}
\\
MetaTrain-13&{Car Models}&
Classify cars by model&
3229 / 405&
45&
3&
\href{https://kaggle.com/c/car-classificationproject-vision}{Link}
\\
MetaTrain-14&{Cassava Leaf Disease}&
Identify type of leaf disease&
17115 / 2141&
5&
1&
\href{https://kaggle.com/c/cassava-leaf-disease-classification}{Link}
\\
MetaTrain-15&{Celebrity Images}&
Classify celebrity images by attractiveness&
161985 / 20248&
2&
1&
\href{https://kaggle.com/jessicali9530/celeba-dataset}{Link}
\\
MetaTrain-16&{Chess Pieces}&
Identify chess pieces&
437 / 54&
6&
1&
\href{https://kaggle.com/niteshfre/chessman-image-dataset}{Link}
\\
MetaTrain-17&Russian Handwritten Letters&
Classify Russian handwritten letters&
11350 / 1419&
33&
2&
\href{https://kaggle.com/olgabelitskaya/classification-of-handwritten-letters}{Link}
\\
MetaTrain-18&{CT Images}&
Identify intracranial hemorrhage in CT scans&
4255 / 532&
2&
1&
\href{https://kaggle.com/vbookshelf/computed-tomography-ct-images}{Link}
\\
MetaTrain-19&{Corals}&
Identify types of coral&
489 / 62&
14&
1&
\href{https://kaggle.com/c/corales}{Link}
\\
MetaTrain-20&{Cracks}&
Detect cracks in pavements and walls&
13570 / 1697&
2&
1&
\href{https://kaggle.com/c/crack-identification-ce784a-2020-iitk}{Link}
\\
MetaTrain-21&{Cactus Identification}&
Identify cactus in images&
17199 / 2150&
2&
1&
\href{https://kaggle.com/c/cs4487-2020fall}{Link}
\\
MetaTrain-22&{Animals}&
Classify animal pictures by species&
320 / 32&
16&
1&
\href{https://kaggle.com/c/cs4670spring2020pa3}{Link}
\\
MetaTrain-23&{Blink}&
Identify which eye is closed&
3874 / 484&
5&
1&
\href{https://kaggle.com/c/csep546-aut19-kc2}{Link}
\\
MetaTrain-24&{Dogs}&
Classify breeds of dogs&
12558 / 1571&
120&
6&
\href{https://kaggle.com/c/cv2020-classification-challenge}{Link}
\\
MetaTrain-25&{Furniture}&
Identify type of furniture&
5186 / 648&
5&
1&
\href{https://kaggle.com/c/day-3-kaggle-competition}{Link}
\\
MetaTrain-26&{Forest Fire}&
Detect whether there is a fire in forest images&
794 / 100&
3&
1&
\href{https://kaggle.com/c/defi1-ia}{Link}
\\
MetaTrain-27&{Devanagari Characters}&
Identify Devanagari characters&
73580 / 9197&
46&
3&
\href{https://kaggle.com/rishianand/devanagari-character-set}{Link}
\\
MetaTrain-28&{COVID Chest X-Ray}&
Identify COVID by chest x-ray pictures&
599 / 75&
2&
1&
\href{https://kaggle.com/c/dlai3}{Link}
\\
MetaTrain-29&{Bottles}&
Identify how full a soda bottle is&
11992 / 1499&
5&
1&
\href{https://kaggle.com/c/e4040fall2019-assignment-2-task-5}{Link}
\\
MetaTrain-30&{Indoor Scenes}&
Identify the kind of indoor place&
2498 / 312&
10&
1&
\href{https://kaggle.com/c/fcis-sc-deeplearning-competition}{Link}
\\
MetaTrain-31&{Flowers}&
Recognize flower types&
3455 / 431&
5&
1&
\href{https://kaggle.com/alxmamaev/flowers-recognition}{Link}
\\
MetaTrain-32&{Four Shapes}&
Identify basic shapes&
11976 / 1496&
4&
1&
\href{https://kaggle.com/smeschke/four-shapes}{Link}
\\
MetaTrain-33&{Fruits}&
Identify fruits in different lighting conditions&
35091 / 4386&
15&
1&
\href{https://kaggle.com/chrisfilo/fruit-recognition}{Link}
\\
MetaTrain-34&{Fruits 360}&
Identify fruits in various orientations&
72225 / 9057&
131&
7&
\href{https://kaggle.com/moltean/fruits}{Link}
\\
MetaTrain-35&{Garbage}&
Classify garbage types&
2019 / 252&
6&
1&
\href{https://kaggle.com/asdasdasasdas/garbage-classification}{Link}
\\
MetaTrain-36&{Handwritten digits}&
Identify handwritten digits&
47995 / 5999&
10&
1&
\href{https://kaggle.com/c/gen-2-ai-force-challenge-1}{Link}
\\
MetaTrain-37&{Emojis}&
Identify the type of emojis from various styles&
5324 / 667&
50&
3&
\href{https://kaggle.com/c/gpa759-2020}{Link}
\\
MetaTrain-38&{German Traffic Signs}&
Classify german traffic signs&
31367 / 3921&
43&
3&
\href{https://kaggle.com/meowmeowmeowmeowmeow/gtsrb-german-traffic-sign}{Link}
\\
MetaTrain-39&{Flowers 2}&
Identify type of flower&
5194 / 651&
102&
6&
\href{https://kaggle.com/spaics/hackathon-blossom-flower-classification}{Link}
\\
MetaTrain-40&{Scraped Images}&
Classify web images into four generic categories&
27258 / 3408&
4&
1&
\href{https://kaggle.com/duttadebadri/image-classification}{Link}
\\
MetaTrain-41&{Natural Images}&
Classify natural images into six generic categories&
13620 / 1703&
6&
1&
\href{https://kaggle.com/puneet6060/intel-image-classification}{Link}
\\
MetaTrain-42&{Animals and Objects}&
Identify images of objects and animals&
4000 / 500&
10&
1&
\href{https://kaggle.com/c/khu-deep-learning-competition}{Link}
\\
MetaTrain-43&{Ships}&
Identify types of ships&
3998 / 500&
5&
1&
\href{https://kaggle.com/c/kunstmatigeintelligentie20192020}{Link}
\\
MetaTrain-44&{Surgical Tools}&
Classify surgical tools&
648 / 81&
4&
1&
\href{https://kaggle.com/dilavado/labeled-surgical-tools}{Link}
\\
MetaTrain-45&{Land Use}&
Detect kind of land use from satellite images&
14399 / 1801&
10&
1&
\href{https://kaggle.com/c/land-cover-class}{Link}
\\
MetaTrain-46&{Lego Bricks}&
Classify Lego bricks by shape&
5103 / 638&
16&
1&
\href{https://kaggle.com/joosthazelzet/lego-brick-images}{Link}
\\
MetaTrain-47&{Lego Bricks 2}&
Classify Lego bricks by shape&
7319 / 915&
20&
1&
\href{https://kaggle.com/pacogarciam3/lego-brick-sorting-image-recognition}{Link}
\\
MetaTrain-48&{Lego Minifigures}&
Classify Lego minifigures by franchise&
128 / 14&
14&
1&
\href{https://kaggle.com/ihelon/lego-minifigures-classification}{Link}
\\
MetaTrain-49&{Real or Fake Legos}&
Identify off-brand lego bricks from real ones&
36606 / 4576&
4&
1&
\href{https://kaggle.com/pacogarciam3/lego-vs-generic-brick-image-recognition}{Link}
\\
MetaTrain-50&{Makeup}&
Identify whether a person is wearing makeup&
1203 / 150&
2&
1&
\href{https://kaggle.com/petersunga/make-up-vs-no-make-up}{Link}
\\
MetaTrain-51&{Male Female}&
Identify gender of a person&
46913 / 5864&
2&
1&
\href{https://kaggle.com/c/malefemale-for-drr}{Link}
\\
MetaTrain-52&{Messy vs Clean Room}&
Classify pictures of rooms as either messy or clean&
168 / 22&
2&
1&
\href{https://kaggle.com/cdawn1/messy-vs-clean-room}{Link}
\\
MetaTrain-53&{Cats vs Dogs}&
Identify cats from dogs&
19975 / 2497&
2&
1&
\href{https://kaggle.com/shaunthesheep/microsoft-catsvsdogs-dataset}{Link}
\\
MetaTrain-54&{Flowers 3}&
Identify type of flower&
3109 / 388&
5&
1&
\href{https://kaggle.com/c/mis583-hw2-part-2}{Link}
\\
MetaTrain-55&{Tiny Images}&
Identify type of object from tiny images&
15995 / 2000&
10&
1&
\href{https://kaggle.com/c/mllabgame}{Link}
\\
MetaTrain-56&{Mushroom classification}&
Classify mushrooms by genus&
5312 / 662&
9&
1&
\href{https://kaggle.com/maysee/mushrooms-classification-common-genuss-images}{Link}
\\
MetaTrain-57&{Carpet Anomaly}&
Identify type of anomaly on carpets&
315 / 41&
6&
1&
\href{https://mvtec.com/company/research/datasets/mvtec-ad}{Link}
\\
MetaTrain-58&{Grid Anomaly}&
Identify type of anomaly on a grid&
270 / 33&
6&
1&
\href{https://mvtec.com/company/research/datasets/mvtec-ad}{Link}
\\
MetaTrain-59&{Leather Anomaly}&
Identify type of anomaly on leather&
293 / 38&
6&
1&
\href{https://mvtec.com/company/research/datasets/mvtec-ad}{Link}
\\
MetaTrain-60&{Natural Images 2}&
Identify types of natural images&
11221 / 1402&
6&
1&
\href{https://kaggle.com/c/nnfl-cnn-lab2}{Link}
\\
MetaTrain-61&{Printed Letters}&
Identify printed Latin letters in various fonts&
381052 / 47630&
10&
1&
\href{https://kaggle.com/jwjohnson314/notmnist}{Link}
\\
MetaTrain-62&{Bengali Digits}&
Identify handwritten Bengali digits&
57620 / 7203&
10&
1&
\href{https://kaggle.com/BengaliAI/numta}{Link}
\\
MetaTrain-63&{Flowers 4}&
Identify type of flower&
2855 / 356&
5&
1&
\href{https://kaggle.com/c/nuu-me-midterm-exam-image-classification}{Link}
\\
MetaTrain-64&{Oregon Wildlife}&
Identify type of wildlife in pictures taken in Oregon&
5655 / 708&
20&
1&
\href{https://kaggle.com/virtualdvid/oregon-wildlife}{Link}
\\
MetaTrain-65&{Parkinsons Drawings}&
Identify stage of Parkinson's disease by drawing&
162 / 20&
2&
1&
\href{https://kaggle.com/kmader/parkinsons-drawings}{Link}
\\
MetaTrain-66&{Dogs 2}&
Classify types of dogs&
7216 / 902&
10&
1&
\href{https://kaggle.com/c/perritos}{Link}
\\
MetaTrain-67&{Seedlings}&
Determine type of plant from a picture of its seedling&
3792 / 474&
12&
1&
\href{https://kaggle.com/c/plant-seedlings-classification}{Link}
\\
MetaTrain-68&{Traffic Signs}&
Identify traffic signs&
5735 / 717&
8&
1&
\href{https://kaggle.com/c/proptit-aif-homework-1}{Link}
\\
MetaTrain-69&{Real vs Fake Faces}&
Identify fake face images from real ones&
1632 / 204&
2&
1&
\href{https://kaggle.com/ciplab/real-and-fake-face-detection}{Link}
\\
MetaTrain-70&{Casting Products}&
Idendify defects in products manufactured by casting&
6866 / 858&
2&
1&
\href{https://kaggle.com/ravirajsinh45/real-life-industrial-dataset-of-casting-product}{Link}
\\
MetaTrain-71&{Rock Paper Scissors}&
Identify hand gestures&
1749 / 219&
3&
1&
\href{https://kaggle.com/drgfreeman/rockpaperscissors}{Link}
\\
MetaTrain-72&{Chest X-ray}&
Identify various information from chest x-ray images&
4484 / 560&
2&
1&
\href{https://kaggle.com/nih-chest-xrays/sample}{Link}
\\
MetaTrain-73&{Furniture 2}&
Classify type of furniture&
2400 / 200&
200&
10&
\href{https://kaggle.com/c/sfu-cmpt-computer-vision-course-cnn}{Link}
\\
MetaTrain-74&{Sheep}&
Classify sheep breeds&
1344 / 168&
4&
1&
\href{https://kaggle.com/divyansh22/sheep-breed-classification}{Link}
\\
MetaTrain-75&{Simpsons}&
Identify characters from a popular TV show&
15969 / 1999&
39&
2&
\href{https://kaggle.com/c/simpsons-challenge-gft}{Link}
\\
MetaTrain-76&{Simpsons 2}&
Identify characters from a popular TV show&
16709 / 2090&
39&
2&
\href{https://kaggle.com/c/simpsons4}{Link}
\\
MetaTrain-77&{Skin Cancer}&
Classify type of Skin Cancer&
1785 / 224&
9&
1&
\href{https://kaggle.com/nodoubttome/skin-cancer9-classesisic}{Link}
\\
MetaTrain-78&{Signed Digits}&
Identify sign language digits&
1644 / 208&
10&
1&
\href{https://kaggle.com/c/sldc}{Link}
\\
MetaTrain-79&{Stanford Dogs}&
Identify dog breeds&
16376 / 2050&
120&
6&
\href{https://kaggle.com/jessicali9530/stanford-dogs-dataset}{Link}
\\
MetaTrain-80&{Preprocessed Stanford Dogs}&
Preprocessed version of Stanford Dogs&
16418 / 2052&
120&
6&
\href{https://kaggle.com/miljan/stanford-dogs-dataset-traintest}{Link}
\\
MetaTrain-81&{Synthetic Digits}&
Identify digits on randomly generated backgrounds&
9600 / 1200&
10&
1&
\href{https://kaggle.com/prasunroy/synthetic-digits}{Link}
\\
MetaTrain-82&{Ethiopic Digits}&
Classify Ethiopic Digits&
48000 / 6000&
10&
1&
\href{https://kaggle.com/c/tau-ethiopic-digit-recognition}{Link}
\\
MetaTrain-83&{Simpsons 3}&
Identify characters from a TV show&
16709 / 2090&
39&
2&
\href{https://kaggle.com/alexattia/the-simpsons-characters-dataset}{Link}
\\
MetaTrain-84&{Traffic Signs 2}&
Identify traffic signs&
20288 / 2530&
67&
4&
\href{https://kaggle.com/c/tl-signs-hse-itmo-2020-winter}{Link}
\\
MetaTrain-85&{Vehicles}&
Classify types of auto vehicles&
22427 / 2803&
17&
1&
\href{https://kaggle.com/c/vehicle}{Link}
\\
MetaTrain-86&{Clothes}&
Identify types of clothes&
12935 / 1617&
6&
1&
\href{https://kaggle.com/dqmonn/zalando-store-crawl}{Link}
\\\midrule
MetaTest-1&{Alien vs Predator}&
Tell apart characters from a movie&
711 / 89&
2&-&
\href{https://www.kaggle.com/pmigdal/alien-vs-predator-images}{Link}
\\
MetaTest-2&{Colorectal Histology}&
Classify colorectal tissue images&
4000 / 496&
8&-&
\href{https://www.kaggle.com/kmader/colorectal-histology-mnist}{Link}
\\
MetaTest-3&{COVID-19}&
Identify lung diseases from radiographic images&
2298 / 288&
3&-&
\href{https://www.kaggle.com/tawsifurrahman/covid19-radiography-database}{Link}
\\
MetaTest-4&{Speed Limit Signs}&
Classify road speed limit signs&
272 / 35&
4&-&
\href{https://www.kaggle.com/c/drr-sign}{Link}
\\
MetaTest-5&{Gemstones}&
Classify different kinds of gemstones&
2206 / 278&
18 (87)\daggerfootnote{\label{note1}The original dataset's number of classes are written in parentheses.}&-&
\href{https://www.kaggle.com/lsind18/gemstones-images}{Link}
\\
MetaTest-6&{Honeybee Pollen}&
Detect whether a honeybee is carrying pollen&
571 / 71&
2&-&
\href{https://www.kaggle.com/ivanfel/honey-bee-pollen}{Link}
\\
MetaTest-7&{Chinese Characters}&
Identify handwritten Chinese characters&
13762 / 1715&
20 (200)
&-&
\href{https://www.kaggle.com/anokas/kuzushiji}{Link}
\\
MetaTest-8&{Real or Drawing}&
Identify real images from drawings in tiny images&
4000 / 500&
10&-&
\href{https://www.kaggle.com/c/ml2020spring-hw12}{Link}
\\
MetaTest-9&{Dessert}&
Identify types of dessert&
1324 / 166&
5&-&
\href{https://www.kaggle.com/c/recognizance1}{Link}
\\
MetaTest-10&{Dog Breeds}&
Classify dog breeds&
5295 / 656&
19 (133)
&-&
\href{https://www.kaggle.com/c/ucfai-core-fa19-cnns}{Link}
\\
\end{longtable}

\endgroup

\end{center}

\twocolumn

\end{document}